\documentclass{article}

\usepackage{arxiv}

\usepackage[utf8]{inputenc} 
\usepackage[T1]{fontenc}    
\usepackage{hyperref}       
\usepackage{url}            
\usepackage{booktabs}       
\usepackage{amsfonts}       
\usepackage{nicefrac}       
\usepackage{microtype}      
\usepackage{lipsum}

\usepackage[bottom]{footmisc}

\usepackage{setspace} 
\usepackage[utf8]{inputenc} 
\usepackage{amsmath, amsthm, amssymb}	
\usepackage{amssymb}               
\usepackage{graphicx}              
\usepackage{epstopdf}
\usepackage{commath}
\usepackage{lscape}
\usepackage{amsfonts}
\usepackage{icomma} 
\usepackage[small,bf,hang]{caption}

\usepackage{algorithm}
\usepackage[noend]{algpseudocode}

\usepackage{booktabs}
\usepackage{pgfplots}
\usepackage{pgfplotstable}

\newcommand{\RomanNumeralCaps}[1]
{\MakeUppercase{\romannumeral #1}}

\author{
	Yanou Ramon\thanks{Corresponding author.} \\
	Department of Engineering Management\\
	University of Antwerp\\
	\And
	David Martens \\
	Department of Engineering Management\\
	University of Antwerp\\
	\And
	Foster Provost \\
	Stern School of Business (NYU) \\
	\And
	Theodoros Evgeniou \\
	INSEAD Paris \\
}

\begin{document}
\title{Counterfactual Explanation Algorithms for Behavioral and Textual Data}
\maketitle


\begin{abstract}
We study the interpretability of predictive systems that use high-dimensonal behavioral and textual data. Examples include predicting product interest based on online browsing data and detecting spam emails or objectionable web content. Because these data are very high-dimensional, serious comprehensibility issues arise: non-linear models, which are difficult to understand in the first place, become completely opaque when using thousands of features, and even linear models require the investigation of thousands of coefficients. Behavioral and text data instances also tend to be sparse, which means that any model component may be irrelevant to a particular instance. 
Recently, counterfactual explanations have been proposed for generating insight into model predictions, which 
bypass issues of model opaqueness and coefficient interpretation, and focus on what is relevant to a particular instance. Conducting a complete search to compute counterfactuals is very time-consuming because of the huge dimensionality. To our knowledge, for behavioral and text data, only one model-agnostic heuristic algorithm (\textit{SEDC}) for finding counterfactual explanations (``Evidence Counterfactuals``) has been proposed in the literature. However, there may be better algorithms for finding counterfactuals quickly. This study aligns the recently proposed Linear Interpretable Model-agnostic Explainer (LIME) and Shapley Additive Explanations (SHAP) with the notion of counterfactual explanations, and empirically benchmarks their effectiveness and efficiency against \textit{SEDC} using a collection of $13$ data sets. Results show that LIME-Counterfactual \textit{(LIME-C)} and SHAP-Counterfactual \textit{(SHAP-C)} have low and stable computation times, but mostly, they are less efficient than \textit{SEDC}. However, for certain instances on certain data sets, \textit{SEDC}'s run time is comparably large. With regard to effectiveness, \textit{LIME-C} and \textit{SHAP-C} find reasonable, if not always optimal, counterfactual explanations. \textit{SHAP-C}, however, seems to have difficulties with highly unbalanced data. Because of its good overall performance, \textit{LIME-C } seems to be a favorable alternative to \textit{SEDC}, which failed for some nonlinear models to find counterfactuals because of the particular heuristic search algorithm it uses. A main upshot of this paper is that there is a good deal of room for further research. For example, we propose algorithmic adjustments 
that are direct upshots of the paper's findings.
\end{abstract}

\section{Introduction}
\label{Section:Introduction}

The proliferation of big data architectures has resulted in predictive modeling applications having an increasingly large impact on business and society~\cite{2014JunqueBigData}. We focus on two sorts of big data. The first is behavioral data, defined as data that capture human behavior through the actions and interactions of people~\cite{2017Shmueli}, which can be used for various predictive purposes. For instance, digital records of behavior such as Facebook `Like' data~\cite{2013KosinskiFacebook,2017Kosinski}, Twitter profiles~\cite{2011KosinskiTwitter} or music collections~\cite{2018NaveKosinskiMusic} can be used to infer psychological traits and political orientation~\cite{2013KosinskiFacebook,2017Kosinski,2018PraetFacebook}, while the merchants you pay to or webpages you visit are predictive for product interest~\cite{2016MartensMassive} and creditworthiness~\cite{2015DeCnuddeMicrofinance}. The second big data source is textual data, which often is preprocessed into a similar high-dimensional, sparse representation for predictive modeling. Text classification is ubiquitous in business and government.

Mining behavior and text can result in highly accurate classification models~\cite{2014JunqueBigData,2016MartensMassive}, but also in very complex model structures. The complexity arises from either the learning technique (e.g., deep learning) or the data, or both. Behavioral and textual data are typically high-dimensional and sparse. Let's consider an example, that we will refer back to.  We want to predict the gender of users based on the movies they have viewed. A user having watched a movie is represented by a binary feature for each movie, which results in an enormous feature set. However, each user only has watched a small number of movies, which results in an extremely sparse data matrix (almost all elements are zero). 
Because of these data characteristics, even normally interpretable linear models are difficult to interpret because there are many thousands of features, each with their own linear coefficient; further, the features that will be brought to bear for prediction are different for every individual. 
Moreover, applying nonlinear techniques normally renders the reasons for a particular prediction completely opaque. 


The importance of understanding classification decisions is well-argued in the literature~\cite{2018RasExplanationsDL,2016LiptonMythosInterpret,2014FreitasComprehensibleModels,1999BenbasatExplanations,2017DoshiKimInterpretableML,2014MartensSEDC}.
Explanations for model predictions are often necessary for users to trust, accept and improve the decision system~\cite{1999BenbasatExplanations}. In some domains, like medical diagnosis and credit scoring, it even is a legal requirement~\cite{2014MartensSEDC,1999BenbasatExplanations,2007MartensRulesEJOR} (e.g., \textit{why was my loan application rejected?}).
According to Doshi-Velez and Kim~\cite{2017DoshiKimInterpretableML}, the demand for interpretable models stems from a mismatch between formal objectives (e.g., minimize the prediction error) and ethical objectives (e.g., privacy). The latter can only be validated when a certain level of interpretability is achieved. Moreover, explanations can reveal overfitting by the model or other issues such as data leakage, which may not always be revealed by evaluation criteria such as the area under the ROC curve~\cite{2014MartensSEDC,2016RibeiroLIME}.

\begin{figure*}
	\centering
	\includegraphics[width=0.8\textwidth, clip]{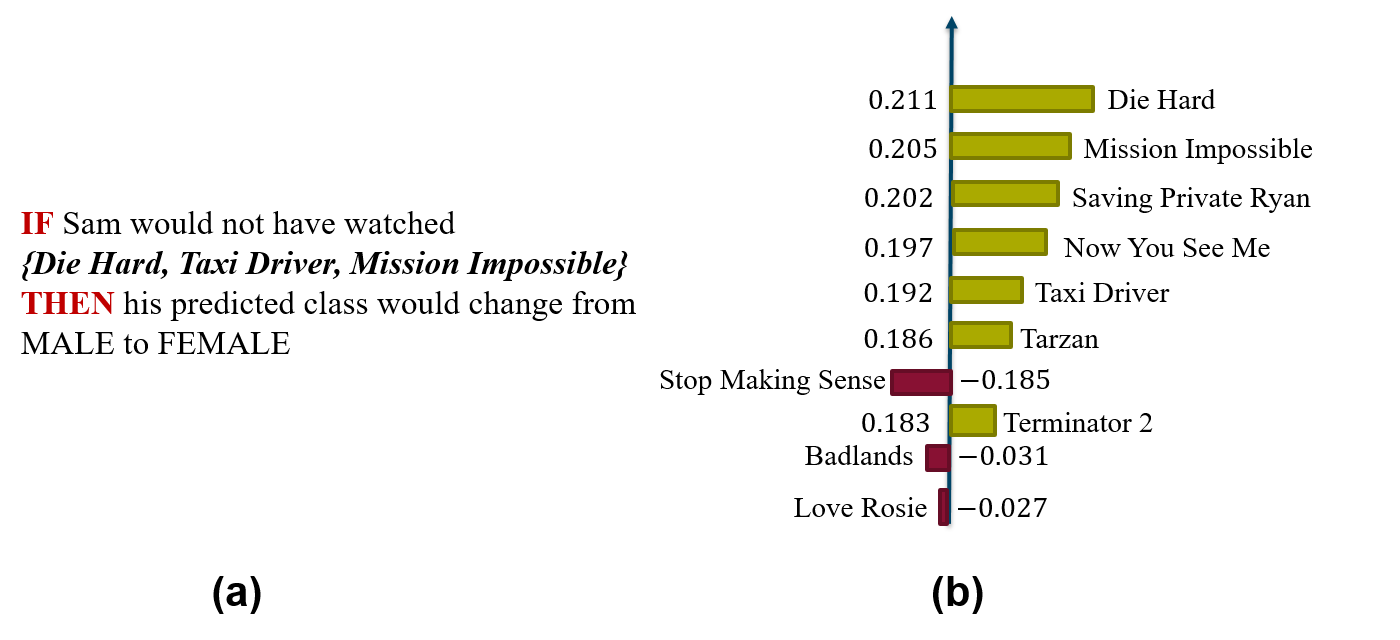}
	\caption{(a) Counterfactual explanation and (b) additive feature attribution explanation.}
	\label{edcexample}
\end{figure*}

Various approaches have been proposed for explaining model predictions, varying in scope, flexibility and output type~\cite{2016LiptonMythosInterpret,2014MartensSEDC,2016RibeiroLIME,2017LundbergLeeSHAP,2017WachterCounterfactuals}. 
For this paper, we focus on the increasingly popular notion of `counterfactual explanations'~\cite{2019FlachCounterfactuals,2017WachterCounterfactuals,2018NguyenText,2017ChenCloaking,ProvostKeynote,2014MartensSEDC}\footnote{We will refer to counterfactual explanations as counterfactuals, explanation subsets or explanations interchangeably.}. 
A counterfactual explanation of a model-based system's decision for a particular instance comprises a set of feature values of the instance without which the system would not have made that decision. In our running movie example, if we want an explanation of why user Sam was predicted to be `male', we want to know which movies were critical for the model's decision. A counterfactual explanation shows a set of movies such that removing them (~setting feature values to zero) from Sam's movie list would lead the predicted class to no longer be `male' (see \textbf{Figure~\ref{edcexample}}\textbf{a}). 

In this study, we are interested in finding the minimum-sized counterfactual explanation. One possible approach to find counterfactual explanations is to conduct a complete search through the entire space of feature combinations.  If we wanted, say, to find the smallest counterfactual explanation, we could start with one feature and incrementally increasing the number of features until an explanation is found. However, this strategy scales exponentially with the number of features, making it impracticable and possibly ineffective for high-dimensional feature spaces~\cite{2014MartensSEDC} 
and effectiveness.

Martens and Provost~\cite{2014MartensSEDC} proposed a heuristic best-first search for finding Evidence Counterfactuals (\textit{SEDC})\footnote{The original paper presented the framework for counterfactual explanations, subsequently referred to as Evidence Counterfactuals~\cite{ProvostKeynote,2016MoeyersomsGlobalExp,2017ChenCloaking}. The paper presented a model-agnostic method for finding such explanations and a method for linear models. We evaluate the heuristic best-first search \textit{SEDC} algorithm here.}, which is able to counterfactually explain predictions of any classification model. To our knowledge, this has been the only proposed model-agnostic algorithm for counterfactuals which is able to deal with behavioral and textual data sources. 

In the literature, other instance-level explanation types have been proposed for very high-dimensional data sources, such as additive feature attribution explanations~\cite{2017LundbergLeeSHAP,2016RibeiroLIME}. 
In our movie running example, such explanations would show an ordered list of movies and their corresponding importance weights -- specifically, importance for this particular model decision (see \textbf{Figure~\ref{edcexample}}\textbf{b}). 
The idea of developing hybrid methods which connect counterfactuals with additive feature attribution explanations stems from the following reasoning: if these importance-rankings of features are sufficiently accurate, it may be possible to compute counterfactuals from them: starting from the highest-ranked feature, we remove features until the predicted class changes. 
The novelty of this study resides in the idea that these importance rankings may be an `intelligent' starting point for searching for counterfactuals. The resulting algorithm for computing counterfactuals may be better than the existing \textit{SEDC} algorithm. For this reason, we empirically compare the counterfactual explanation algorithms to help researchers and practitioners better understand which method is most suitable when facing behavioral or textual data.

This paper's main contributions are fourfold: (1) we propose two novel model-agnostic explanation algorithms, creating them via the combination of the notion of counterfactuals and additive feature attribution methods (\textit{LIME-C} and \textit{SHAP-C}); (2) we define quantitative evaluation criteria that proxy for the effectiveness and efficiency of these algorithms; (3) we perform an in-depth evaluation of the explanation quality of \textit{LIME-C} and \textit{SHAP-C} when applied to high-dimensional behavioral and textual data and benchmark their performance against the existing \textit{SEDC} algorithm, and lastly, (4) we highlight that this is an area ripe for additional research.  For example, based directly on the paper's main findings, we propose changes to the model-agnostic methods for generating counterfactuals.

\vspace*{-5mm}
\section{Couterfactual Explanations}
\label{sec:1}
\vspace*{-2mm}

\paragraph{Motivation.}
Much recent research on `Explainable AI' and `Interpretable Machine Learning' focuses on making predictive models more interpretable. To achieve interpretability of classification models, there are two main approaches: (1) restrict oursleves to inherently interpretable models or (2) post-process our models in ways that yield insights into how our model works~\cite{2019MurdochIML}. Model-based explanation methods naturally embed an explanation in the model due to its simplicity~\cite{2017DoshiKimInterpretableML,2014FreitasComprehensibleModels}.  
As model-focused interpretation largely limits the choice of models to linear models, rule-based models, and decision trees, they may result in lower predictive performance on complex modeling problems. In contrast, complicated, nonlinear models can yield high predictive accuracy, but are harder to analyze, which results in lower interpretability. Moreover, in the context of high-dimensional, sparse data, even the use of intrinsically interpretable models such as linear models requires inspecting all the learned weights, making model-focused explanation also not useful in practice for example for high-dimensional data. 

For this reason, post-hoc methods that use the model's inputs and outputs to increase understanding, may be more helpful. 
Many post-hoc interpretation approaches have been proposed, varying in scope, flexibility and output type. The scope regards whether the method generates global explanations (for the entire feature and instance space) or instance-level explanations (for a single model prediction), whereas the flexibility regards whether the approach is model-specific or model-agnostic. Much research focuses on model-specific explanation techniques tailored to a specific type of predictive models such as deep learning models~\cite{2016ArrasLRPText,2015SamekVisualizationDL} or random forests~\cite{2001BreimanRF}. In contrast, model-agnostic methods explain predictions of any predictive model. This increases flexibility; however, often it results in substantially more computational effort~\cite{2016ArrasLRPText,2014MartensSEDC}. Lastly, the proposed methods can have different types of explanation such as a (set of) rule(s)~\cite{2014MartensSEDC,2017WachterCounterfactuals,2018GuidottiLORE}, feature importance rankings~\cite{2017LundbergLeeSHAP,2016RibeiroLIME,2016ArrasLRPText,2001BreimanRF} or visual explanations~\cite{2016ArrasLRPText,2015SamekVisualizationDL}. 
Another important class of explanations, on which we focus in this study, are counterfactual explanations that seek minimal changes to the feature values such that the model's predicted outcome changes. 
To our knowledge, counterfactual explanations for classification decisions were first used to explain document classifications~\cite{2014MartensSEDC}, and since have been applied more broadly~\cite{2019FlachCounterfactuals,2017WachterCounterfactuals,2018NguyenText,2017ChenCloaking,ProvostKeynote,2014MartensSEDC}. Martens \& Provost \cite{2014MartensSEDC} define a (counterfactual) explanation as a minimal (irreducible) set of features such that, when removing these features from the instance, the predicted class changes.\footnote{Such explanations have been called `Evidence Counterfactuals', referring to the feature evidence that leads the classifier to make its classification~\cite{ProvostKeynote,2017ChenCloaking}; we will adopt this terminology to differentiate such explanations from the additive feature explanations described next.}  
For instance, in \textbf{Figure~\ref{edcexample}}\textbf{a}, the movies \textit{``Die Hard"}, \textit{``Taxi Driver"} and \textit{``Mission Impossible"} explain why Sam was classified as `male'.
Counterfactuals are comprehensible because of their concise and easily-understandable format (i.e., a rule)~\cite{2019FlachCounterfactuals}. Moreover, Wachter et al.~\cite{2017WachterCounterfactuals} argue that counterfactuals are the first step towards balancing model comprehensibility with other interests such as minimizing regulatory burden on businesses or preserving other data subjects' privacy, while also increasing public acceptance of automatic decision making. Counterfactuals have been argued to be crucial for explaining predictions on the instance-level as they pinpoint the features that led to each specific decision~\cite{2019FlachCounterfactuals} and can make the decision actionable (e.g., \textit{what change my data would lead to a different outcome?})~\cite{2017WachterCounterfactuals,2017ChenCloaking,2019RusselCounterfactuals}.
The use of counterfactuals has also received substantial support from the social sciences and philosophical literature~\cite{2018MillerSocialSciences,1973LewisCounterfactuals}, that has studied how people define, generate, select and present explanations, and hence, is a good starting point for explanations in artificial intelligence. Lewis~\cite{1973LewisCounterfactuals} defines a counterfactual as a close possible world in which a different outcome (or model's prediction) occurs. 
Counterfactual explanations are an answer to a \textit{why}-question~\cite{2018MillerSocialSciences}, more specifically, they answer a contrastive question (e.g., \textit{why X rather than not-X?}). Such explanations are argued to be most intuitive and valuable for humans~\cite{2018MillerSocialSciences,2019FlachCounterfactuals,2017WachterCounterfactuals}. According to Miller's~\cite{2018MillerSocialSciences} overview of findings in social sciences and philosophical research on explanations, 
explanations should focus on counterfactuals and state the affairs that would have resulted from some event that did not occur. 

Given our focus on ultra-high-dimensional, ultra-sparse data, in this paper we will consider counterfactuals based on the removal of evidence that is present in the data---for example, words that appear in the document or items that an individual has Liked on Facebook.  These correspond to ``active'' features---those that are present in a sparse representation, or those that are non-zero in a traditional feature-vector representation.  The interested reader is referred to Martens \& Provost~\cite{2014MartensSEDC} for further discussion of this choice and alternatives, for example in the case where a feature not being present would be considered significant evidence for a particular class.

\paragraph{Generating counterfactual explanations.}

As argued by Martens and Provost~\cite{2014MartensSEDC}, the objectives of a search algorithm can vary greatly as a user may want to find (at least) one minimum-sized explanation, find all counterfactual explanations, find as many explanations as possible within a fixed time period or find one explanation as quickly as possible. In this study, we focus on the former objective: 
we wish to find a minimum-sized counterfactual explanation to explain a (positively-predicted) instance $x$ = $(x_{1},...,x_{m})$, where $m$ represents the number of features. We define the instance $x'$ $\in$ $\{0,1\}^{m}$ as the binary representation of the original $\textbf{x}$ (nonzero or \textit{active} value represents $1$, else $0$). A mapping function $h(\cdot)$ is used to transform the original feature vector $x$ to the binary representation $x'$ (nonzero values of $x$ are transformed to a $1$). \textbf{Figure~\ref{fig:representation}} shows more details about the mapping from the original representation to the simplified, binary representation in the context of behavioral and textual data.

We assume that a (trained) binary classifier $C_{M}$ is given that maps the feature vectors or instances $x$ to a binary space ($1$ is positively predicted, $0$ if negatively predicted). The expression $C_{M}(x)$ = $1$ indicates that the instance $x$ is predicted as a positive or as the class-of-interest. This classifier has a corresponding scoring function $f_{C_{M}}$. This scoring function  together with a specific threshold value $t$ enable the classifier $C_{M}$ to turn the predicted scores into explicit predictions ($1$s and $0$s). The predicted scores which lie above this $t$-value are transformed to a positively-predicted label $1$; otherwise, the predicted label is $0$ (negatively-predicted instance). The predicted score of the positively-predicted instance $x$ for which we want to get an explanation thus lies above the threshold value: $f_{C_{M}}(x)$$\geq$$t$.
We define a perturbed instance $z$ as an instance derived from the original instance $x$ that replaces a subset of all active features of instance $x$ by $0$s. All feature values $x_{k}$ of the instance $x$ that are zero, remain zero in the perturbed instance $z$. In other words, only the active feature values of $x$ can be ``perturbed'' or replaced by a zero value. Given a set of indices $I$ which forms a subset of the set of indices of active features $I_{A}$ of the instance $x$ ($I$ $\subseteq $ $I_{A}$), we can define a perturbed instance $z$=$(z_{1},...,z_{m})$ as:

\begin{equation}
z_{I} = z = \left\{
		\begin{array}{ll}
		\forall j \in I: z_{j}=0\\
		\forall j \notin I: z_{j}=x_{j}\\
		\end{array}
\right. 
\label{eq_perturbed}
\end{equation}

A counterfactual explanation $R$ shows a set of (active) features of instance $x$ such that setting these feature values to zero, the predicted class changes. The corresponding indices of the features in $R$ are defined by the set $I_{R}$. 
There exists a perturbed instance $z_{I_{R}}$ that is derived from the original instance $x$ and does no longer have nonzero values for the features in $R$ with corresponding indices $I_{R}$. In other words, the features in the explanation $R$ are nonzero in the original feature $x$, but have a corresponding zero feature value in the perturbed instance $z_{I_{R}}$. If we are looking for an optimal counterfactual explanation $R^{\ast}$, we are, in fact, also looking for an optimal perturbed instance $z^{\ast}_{I_{R^{\ast}}}$ (or short $z^{\ast}$). The minimum-sized counterfactual explanation $R^{\ast}$ is equivalent to the perturbed instance $z^{\ast}$ that flips the fewest possible active feature values of $x$ to zero. 
\\
\\
We optimize equations~\ref{eq_objective1} and~\ref{eq_objective2} using the lexicographical method so to find the minimum-sized counterfactual explanation $R^{\ast}$ (or equivalently, the perturbed instance $z^{\ast}$ that is as close to the original instance $x$ as possible and has a different predicted class): 
\begin{equation}
\begin{aligned}
A=\{ \mbox{z} | (\mbox{z}=\underset{z}{\mathrm{argmin}} \mbox{ d(z,x)}) \land (\mbox{f(z)} < t) \land (\forall x_{j}>0: z_{j} \in \{0,x_{j}\}) \land (\forall x_{k}=0: z_{k}=0) \}
\label{eq_objective1}
\end{aligned}
\end{equation}

\begin{equation}
z^{\ast} = \underset{z \in A}{\mathrm{argmin}} \mbox{ f(z)}
\label{eq_objective2}
\end{equation}

A counterfactual explanation $R^{\ast}$ resulting from the above equations is the smallest-possible set of active features of $x$ that change the predicted class. The instance $z^{\ast}$ corresponds to the closest-possible instance to $x$ that has a different predicted class. 

In equation~\ref{eq_objective1}, $d(\cdot,\cdot)$ measures how far the instance $z$ and the original instance $x$ are from one another. We use the cosine similarity metric between the binary representations of the two instances to measure the distance. More specifically, we first transform $x$ and $z$ onto a binary representation $x'$ and $z'$ (where $1$s refer to nonzero or active values in the original vector, and $0$s refer to zero values in the original vector). The cosine similarity between two instances (vectors) is defined as $cosine(x, y) = \frac {x \cdot y}{||x|| \cdot ||y||}$. So, taking this all into account, the distance function $d(\cdot,\cdot)$ is defined by the following equation:

\begin{equation}
d(z,x) = cosine(z',x') = \frac {z' \cdot x'}{||z'|| \cdot ||x'||}
\label{eq_distancefunction}
\end{equation}

In other words, the more original feature values of instance $x$ are obtained in the new instance $z$, the closer the instances $x$ and $z$. Alternatively, the more feature values of $x$ that are set to zero to change the predicted class, the farther away the perturbed instance $z$ is from the original instance $x$. The instance $z'$ can be transformed to the original feature representation by mapping all nonzero values of $z'$ onto their original value. Minimizing this distance function is equivalent to minimizing the number of features that are part of the counterfactual explanation. This is the explanation that shows the most important features that have led to the decision, while still being human-understandable because only a small number of features are shown to the user. 

The aim of this optimisation problem is to find the minimum-sized counterfactual explanation that also maximizes the predicted score change. For this problem, we use a lexicographic optimization method, where we order the objective functions according to their importance. First, in equation~\ref{eq_objective1}, we look for a set of perturbed instances $A$ that are as close as possible to the original instance $x$ (condition $1$) for which holds that the predicted score falls below the threshold (so that the predicted class flips) (condition $2$). Also, we only allow the active features of the instance $x$ to be possibly changed (conditions $3$ and $4$). The nonactive features of the original instance $x$ remain unchanged and keep their zero value. In equation~\ref{eq_objective2}, we then select the perturbed instance $z^{\ast}$ for which the predicted score decreases the most relative to the threshold value $t$. So, given a set of perturbed instances $z$ that have equal distances to the original instance $x$ and for which the other conditions in equation~\ref{eq_objective1} hold, we select the perturbed instance $z^{\ast}$ that decreases the predicted score the most. This extra preference criterion to choose the optimal perturbed instance $z^{\ast}$ is, in fact, a proxy for the confidence of the counterfactual explanation. Given an equal number of features in a counterfactual explanation, the contribution per feature to the predicted class of that instance is larger for a counterfactual explanation with larger predicted score change. 


\paragraph{Why complete search fails.}
A straightforward way to find counterfactuals would be to conduct a complete search through the space of all feature combinations, starting with one feature, and increasing the number of features until a subset is found that changes the predicted class. 
There are many possible search orderings a complete search algorithm might take.  For example, a ``simplest-first'' algorithm could start by checking whether removing each active feature from the instance individually would cause a change in the predicted class label. If so, an irreducible subset is found. If the class does not change, the algorithm considers all combinations of features of size $2$, $3$ and so on. For an instance with $m$ active features, without additional knowledge to reduce search, the combination of $k$ features requires $\frac{m!}{(m-k)!k!}$ evaluations. This complete search will scale exponentially with the number of active features $m$ of the instance and the required number of features $k$ in the subset to change the predicted class. 

For high-dimensional data sets like text and behavior, complete search is infeasible as the computation time becomes too large, especially when the instance has a large number of active features or when the number of feature changes required to change the predicted class is large. The explanations found by complete search will be irreducible, implying that only when \textit{all} features in the explanation are removed, the class label changes. Thus, the complete search is guaranteed to find the minimum-sized explanations.

We can illustrate with the following example using the \textit{Movielens1m} data set. We want to explain why a certain user is predicted as `male' based on the movie viewing data.
Let's consider the simplest-first complete search algorithm described above, and consider a pass through all the combinations of a certain number of features as an ``iteration'' of the search algorithm.
When the number of active features of this instance is very small and/or there are very few iterations required to switch the predicted class, then the computation time for a complete search may still be reasonable. If not, then the computation time will increase exponentially with the number of features to include in an explanation.

\begin{figure*}
	\centering
	\includegraphics[width=0.9\textwidth, clip]{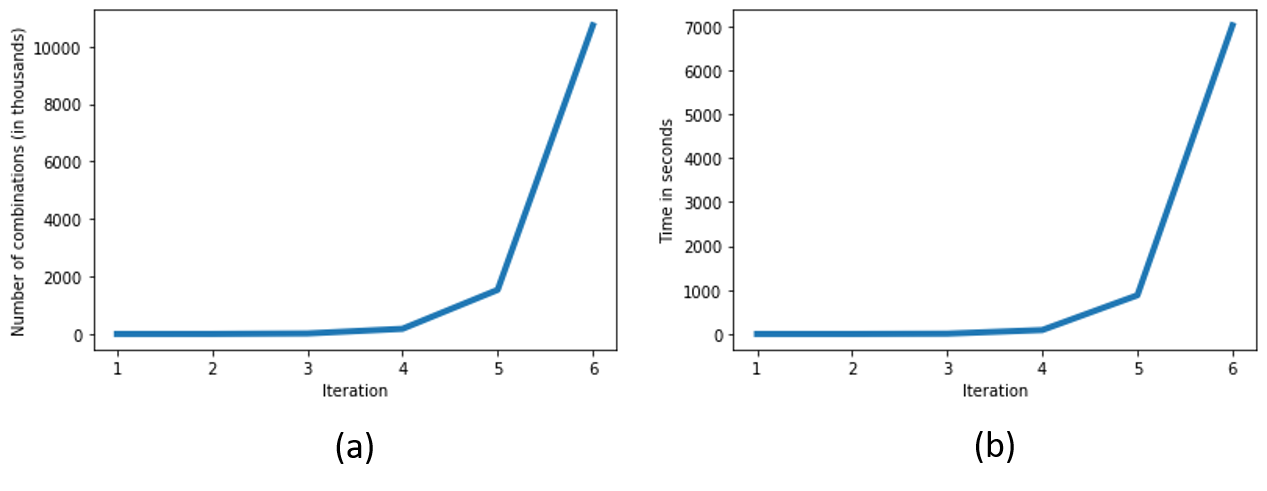}
	\caption{For a complete search, (a) the number of combinations to check (in thousands) and (b) time elapsed (in seconds) in each iteration for a positively predicted instance with $34$ active features and a counterfactual explanation of $6$ features for the \textit{Movielens1m} data and an $l2$-regularized Logistic Regression model.}
	\label{fig:combis_time}
\end{figure*}

\textbf{Figure}\textbf{~\ref{fig:combis_time}} illustrates this. Here, the number of iterations is shown on the $x$-axis, and the number of combinations to check (a) and the time elapsed (b) is shown on the $y$-axis. As the number of iterations increases, the elapsed time and the number of combinations to evaluate, increase non-linearly. The example instance has $34$ active features, meaning this user has watched $34$ movies. To explain why the user was predicted as `male', the counterfactual explanation requires $6$ movies to be removed. So, the complete search algorithm requires $6$ iterations to finally compute this explanation, resulting in a total elapsed time of $116.92$ minutes. It becomes immediately clear that complete search is very time-consuming. For another instance that was predicted to be positive, having $151$ active features, our simple implementation failed to compute a counterfactual because of a memory error after the fourth iteration (when over $20,811,574$ combinations had to be checked). 



\paragraph{Algorithmic choices.}
There is a need for a computationally efficient and effective algorithm for computing counterfactuals in the context of high-dimensional, sparse data. To our knowledge, only one heuristic algorithm has been proposed in the literature, originally used for explaining document classifications~\cite{2014MartensSEDC}. Other proposed algorithms in the literature focus on relatively low-dimensional data with a mixed set of continuous and categorical variables~\cite{2017WachterCounterfactuals, 2019RusselCounterfactuals} or on image data where counterfactuals are computed by means of pixel perturbations~\cite{2018HendricksCounterfactualsImages,2019VanLooverenCounterfactualsPrototypes}. These algorithms cannot deal with many thousands of binary variables---a common representation for explanations for behavior and textual data---which eliminates them from this study.

In this study, we only consider algorithms that are model-agnostic. 
The algorithm should be able to compute counterfactual explanations for model predictions made by \textit{any} classification model. It should only use the input (feature values) and output (model's decision) to come to an explanation. So we look for algorithms that can be applied to any type of model (e.g., linear or nonlinear) and any type of data, irrespective of the size of the data instances, the class imbalance, and so on. 

We make two other algorithmic choices. First of all, we set the maximum size of a counterfactual explanation is $30$ features. That is, we do not allow algorithms to compute counterfactuals that are larger than $30$ features. This is largely for tractability of this research.  However, we note that the utility of very large explanations has been called into question~\cite{2019FlachCounterfactuals,1956MillerCognitiveLimits}. Second, the maximum time to generate a counterfactual explanation is fixed to $2$ minutes. Depending on the application setting, there will be some limit on how long it is reasonable to wait to generate an explanation.  We place a limit that allows us to complete our study---we note that our experience is that if it is going to take more than two minutes, it is generally going to take a lot more than two minutes. 





\subsection{Heuristic Best-First Search}
The only model-agnostic search algorithm for finding counterfactuals for textual and behavioral data has been proposed by Martens \& Provost~\cite{2014MartensSEDC}. We implemented the model-agnostic \textit{SEDC}\footnote{In the original paper, SEDC stands for Search for Explanations for Document Classification~\cite{2014MartensSEDC}. It refers to the model-agnostic search algorithm for computing counterfactual explanations and applies beyond text to other high-dimensional, sparse data.} heuristic search algorithm, which conducts a \textit{best-first} search with local improvement~\cite{2014MartensSEDC}. 
For linear models, \textit{SEDC} is optimal in the sense that it finds the smallest possible feature set (formal proof can be found in~\cite{2014MartensSEDC}). For nonlinear models, because the algorithm cuts off its search after a limit has been reached, optimality is not guaranteed. Also because of the search cut-off, the explanations may not be minimal, i.e. a subset of the explanation set may also be a counterfactual. We further limit \textit{SEDC}'s search by stopping after the first explanation has been found.  As the empirical results below show, this means that the method is quite fast; we leave assessing the full speed vs. effectiveness tradeoff to future work.



The pseudo-code of the algorithm can be found in \textbf{Algorithm~\ref{alg:1}} in \textbf{Appendix~\ref{AppendixA}}. Note that this is a more general version of the pseudo-code in the original paper, as we now also apply it outside the context of document classifications. The algorithm is based on heuristic search guided by local improvement (i.e., best-first search). In a first iteration, the algorithm starts by listing all possible subsets of one feature and calculates the predicted score and class label for each. When a subset of one feature results in a class change, it is added to the list of counterfactual explanations, and the algorithm stops the search. If no subset results in a class change, the algorithm proceeds as a heuristic best-first search. Specifically, in each iteration, the subset for which the predicted output score changes the most in the direction of the opposite class, is expanded with an extra feature. This entails creating a new set of candidate subsets, comprising all combinations with one additional feature from the active features of the instance (that is not yet included in the partial subset). Note that we have added an extra pruning step in the algorithm to make sure that the combinations that were already expanded on, cannot again be expanded on in a new iteration of the algorithm. So, once a feature subset is expanded, this subsets is pruned from the search space in the following iterations. 

\subsection{Novel Hybrid Search Algorithms}
Additive feature attribution methods use an explanation model $g$ as an interpretable approximation of the original model with scoring function $f$ in the neighborhood of an instance. Two recently proposed model-agnostic methods, suitable for high-dimensional data, are the Linear Interpretable Model-agnostic Explainer (LIME)~\cite{2016RibeiroLIME} and Shapley Additive Explanations (SHAP)~\cite{2017LundbergLeeSHAP}. In the context of text and behavior, the explanation model is a linear function of binary variables that indicate whether the feature is present (original value) or absent (zero).
The explanation of an instance $\textbf{x}$ can be represented as,

\begin{equation}
g(\textbf{x}') = \phi_{i0} + \sum_{j=1}^{m} \phi_{ij}x'_{i}
\label{eq:1} 
\end{equation}

where $\textbf{x}'$ $\in$ $\{0,1\}^{m}$ is the binary representation of $\textbf{x}$ (nonzero value represents $1$, else $0$), $m$ is the number of binary features that are nonzero for instance $\textbf{x}$, and $\phi_{i0}, \phi_{ij}$ $\in$ $\mathbb{R}$.
Unlike evidence counterfactual explanations, these explanation models include both positive evidence and negative evidence. For SHAP, the weights retrieved from the model also represent the (approximate) Shapley values, which have theoretically attractive properties~\cite{2017LundbergLeeSHAP}. The main differences between LIME and SHAP are (1) how they generate the sample of perturbed instances (which we refer to as the sampling procedure), (2) the distance function and (3) the complexity control of the explanation, which we describe in the following paragraphs. 

\begin{figure*}
	\centering
	\includegraphics[width=0.9\textwidth, clip]{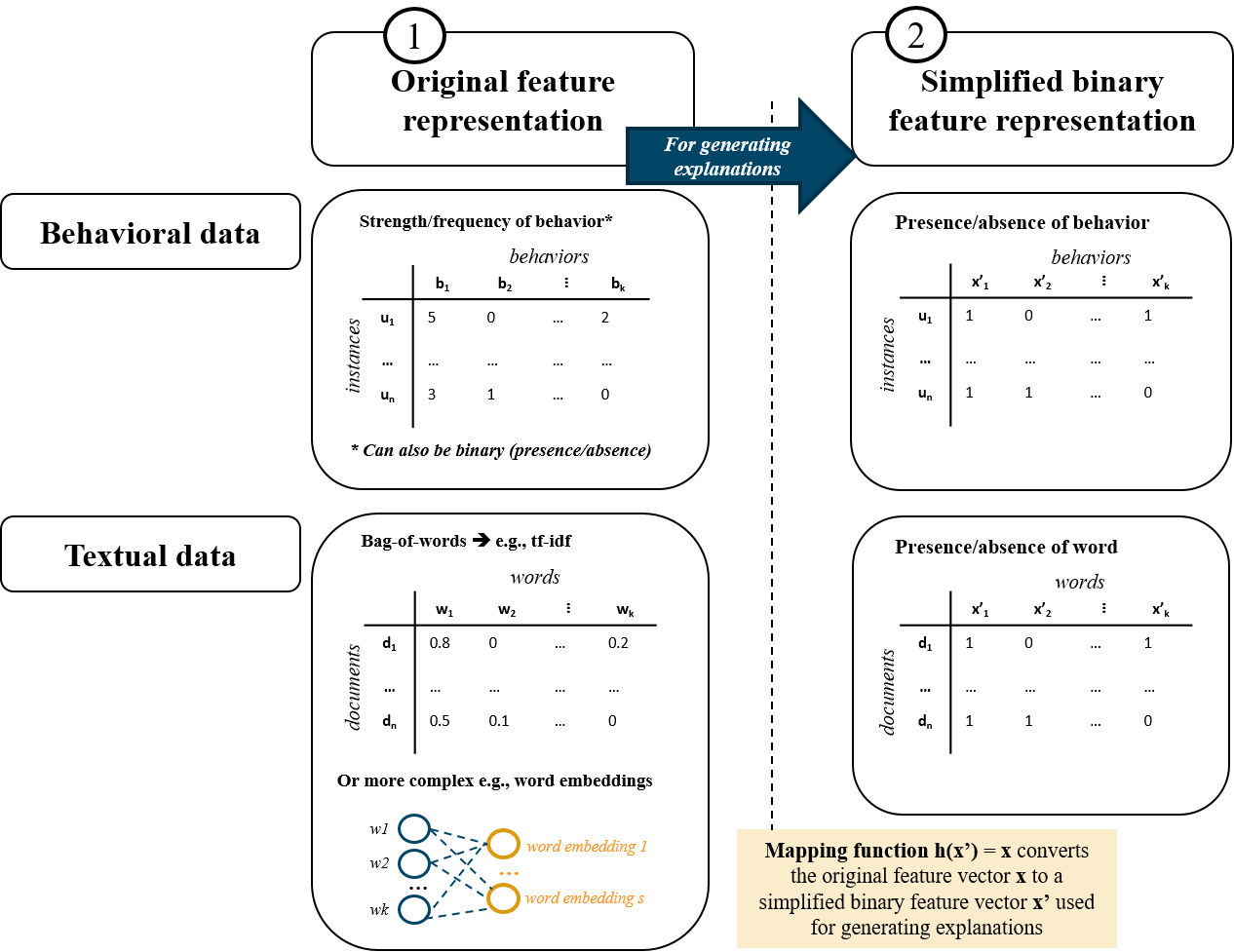}
	\caption{Mapping function $h$ converts the original representation to a simplified, binary representation. From the simplified representation, LIME and SHAP generate perturbed instances.}
	\label{fig:representation}
\end{figure*}

Suppose we want to explain the instance $\textbf{x}_{i}$.
Both LIME and SHAP first map the instance to a binary representation $\textbf{x}'_{i}$=($x'_{i1}$ $\dots$ $x'_{im}$) using a mapping function $h(\textbf{x}_{i}')$=$\textbf{x}_{i}$. (See again \textbf{Figure~\ref{fig:representation}} for more details). 
Next, perturbed instances are generated and each instance $\textbf{z'}$ is assigned a weight $\pi_{\textbf{x}'}(\textbf{z}')$.
LIME generates perturbed samples by sampling $\tilde{n}$ instances by randomly selecting active features of $\textbf{x}'_{i}$. An \emph{active feature} is one whose value is non-zero.
Each perturbed instance $\textbf{z}'$ is then mapped onto the original feature to obtain the predicted score using the original decision function, $f(\textbf{z})$, which is then used as a label for training the explanation model. Each perturbed instance is assigned a corresponding weight. For textual data, LIME uses the sparsity-oriented cosine distance as the distance function to measure the similarity between $\textbf{x}'$ and $\textbf{z}'$, which seems a suitable choice for behavioral data as well.
SHAP starts by estimating distance weights for different subset sizes.
A subset size is the number of nonzero elements of instance $\textbf{z}'$.
For each subset size $s$, a distance weight is estimated. Then, the method samples $\tilde{n}$ perturbed instances from the subset spaces, starting from the smallest (and largest) subsets.

LIME trains the explanation model by using $l2$-regularized linear regression and controls the complexity even more by allowing exactly $K$ features in the explanation.
SHAP trains the model using $l1$-regularized linear regression.

Note that neither LIME nor SHAP produce non-trivial counterfactual explanations natively.  However, it is straighforward to produce variants of the algorithm that do.  Specifically, we can apply the efficient search algorithm for counterfactuals for linear models~\cite{2014MartensSEDC}, which we refer to as \textit{lin-SEDC}\footnote{https://github.com/yramon/edc/tree/master/LinearEDC (see~\cite{2014MartensSEDC} for more details).}, to the importance-ranked lists generated by LIME and SHAP. We refer to these algorithms as \textit{LIME-C} and \textit{SHAP-C} where C stands for ``Counterfactual''. 

The general pseudo-code of a hybrid algorithm of additive feature attribution explanations and counterfactuals is shown in \textbf{Algorithm~\ref{alg:2}} in \textbf{Appendix~\ref{AppendixB}}.
In a first step, an additive feature attribution explanation is generated, without regularizing the explanation model $g$. For LIME, this means that the complexity parameter $K$ is set to the number of active features, whereas for SHAP, the default $l1$-regularization of the author's implementation is used. From this step, a linear model with the binary representation of the features (original value versus zero) is obtained, or equivalently, we retrieve an importance-ranked list of features.

In a second step, the linear algorithm for finding evidence counterfactuals (\textit{lin-SEDC})~\cite{2014MartensSEDC} is applied to the ranked list to efficiently generate a counterfactual smaller than $30$ features, if possible. In more detail: the (active) features of the linear explanation model are ranked by their estimated coefficients. Then, in a first iteration, the feature that is ranked at the top is removed from the instance, or equivalently, its value is set to zero. If this results in a class change, a counterfactual explanation is found. If not, the set of two top-ranked features is checked for being a counterfactual explanation. If not, the set of three top-ranked features are removed from the instance, and so on, until a counterfactual explanation has been found.  As discussed above, we set the maximum size of the counterfactual explanation to $30$, in line with questions as to the utility of explanations sets that are too large~\cite{2019FlachCounterfactuals,2014MartensSEDC}. Both \textit{LIME-C} and \textit{SHAP-C} rely on random sampling to generate counterfactuals, and thus, are stochastic explanation algorithms. This is in contrast to \textit{SEDC}, which always results in the same search tree path for finding explanations when re-running the algorithm. As with our implementation of \textit{SEDC}, there is no guarantee that the counterfactuals from the hybrid algorithms are always minimal.


\vspace*{-5mm}
\section{Experimental setup}
\label{sec:1}
\vspace*{-2mm}

\subsection{Data sets and Models}
Our experimental data consists of $10$ behavioral and $3$ textual data sets. All data are public, except the \textit{Facebook} and \textit{Fraud} data. For the behavioral data sets, we rely on the large collection of data sets for classification tasks provided by~\cite{2015StankovaBipartiteGraphs}. The classification tasks vary from gender prediction to sentiment analysis. 
\textbf{Table~\ref{table:datasets}} summarizes the characteristics of the data. All data have a high-dimensional feature space up to hundreds of thousands of features. \textit{Movielens\_1m, Movielens\_100k, KDD2015, Airline} and \textit{Twitter} have lower-dimensional feature spaces compared to the other data sets.
For all data sets, the class-of-interest is the minority class. A large imbalance is present for the \textit{Fraud} data. Also, \textit{20news} has a large imbalance compared to the other data. Relatively balanced data are \textit{Facebook}, \textit{TaFeng} and \textit{Libimseti} ($b$ larger than $30$\%). The large sparsity values $p$ for all data indicate that the number of active features is very small compared to the total number of possible active features.

\textbf{Table~\ref{table:datasets}} also shows the test instances per data set.
It is interesting to compare the average number of active features, which is very different between the data sets.
\textit{Ecommerce} and \textit{Flickr} have very small instances (only $2$ to $3$ active features), in contrast to other data such as \textit{Movielens\_1m} with instances with over $150$ active features. Note that the table is sorted by increasing values of average number of active features $\dot{m}_{lin}$. 

\begin{table}
	\centering
	\caption{\textbf{Data characteristics} of the data sets: data type (B:behavioral, T:textual), target variable, number of instances and features, imbalance $b$ of the target, the sparsity $p$ and the test set size (percentage of instances predicted as positive are placed in brackets). We use $20$\% of the data as test set. A * indicates that the number of positively predicted test instances used for the experiments was a random subset of 300. The average number of active features $\dot{m}_{lin}$ and $\dot{m}_{nonlin}$ are measured over the positively predicted test instances of respectively the linear and nonlinear model. The last column shows the reference. Note that we sort the data sets by increasing values of $\dot{m}_{lin}$.}\label{table:datasets}
	\scalebox{0.75}{
		\begin{tabular}{cccrrrrrrrr}
			\textbf{Dataset} & \textbf{Type} & \textbf{Target} & \textbf{Instances} &\textbf{Features} & \textbf{$b$} & \textbf{$p$} & \textbf{Test set (\%)} &\textbf{$\dot{m}_{lin}$} &\textbf{$\dot{m}_{nonlin}$} & \textbf{ref} \\
			\noalign{\smallskip}\hline\noalign{\smallskip}
			\\
			Flickr* & B & comments & $100,000$ & $190,991$ & $36.91$\% & $99.99$\% & $20,000$ ($20$\%) & $2.02$ & $2.96$ & ~\cite{flickrDATA}\\
			Ecommerce* & B & gender & $15,000$ & $21,880$ & $21.98$\% & $99.99$\% & $3,000$ ($15$\%) & $2.60$ & $2.67$ & ~\cite{ecommerceDATA}\\
			Airline* & T & sentiment & $14,640$ & $5,183$ & $16.14$\% & $99.82$\% & $2,928$ ($15$\%) & $7.81$ & $8.21$ & ~\cite{airlineDATA}\\
			Twitter & T & topic & $6,090$ & $4,569$ & $9.15$\% & $99.74$\%& $1,218$ ($10$\%) & $9.52$ & $9.35$ & ~\cite{twitterDATA}\\ 
			Fraud* & B & fraudulent & $858,131$ & $107,345$ & $6.4e$-$5$\% & $99.99$\% & $171,627$($1$\%) & $11.83$ & $14.09$ & n.a.\\
			YahooMovies* & B & gender & $7,642$ & $11,915$ & $28.87$\% & $99.76$\% & $1,529$ ($20$\%) & $25.24$ & $25.00$ & ~\cite{yahoomoviesDATA}\\
			TaFeng* & B & age & $31,640$ & $23,719$ & $45.23$\% & $99.90$\% & $6,328$ ($15$\%) & $44.32$ & $37.24$ & ~\cite{huangTaFengDATA}\\
			KDD2015* & B & dropout & $120,542$ & $4,835$ & $20.71$\% & $99.67$\% & $24,109$ ($20$\%) & $49.01$ & $46.40$ & ~\cite{kdd2015DATA}\\
			20news & T & atheism & $18,846$ & $41,356$ & $4.24$\% & $99.84$\% & $3,770$ ($5$\%) & $67.96$ & $62.77$ & ~\cite{20newsDATA}\\
			Movielens\_100k & B & gender & $943$ & $1,682$ & $28.95$\% & $93.69$\% & $189$ ($25$\%) & $68.73$ & $73.42$ & ~\cite{movielensDATA}\\
			Facebook* & B & gender & $386,321$ & $122,924$ & $44.57$\% & $99.94$\% & $77,265$ ($30$\%) & $83.03$ & $84.55$ & ~\cite{2017ChenCloaking} \\
			Movielens\_1m* & B & gender & $6,040$ & $3,706$ & $28.29$\% & $95.53$\%  & $1,208$ ($25$\%) & $168.46$ & $153.46$ & ~\cite{movielensDATA}\\
			Libimseti* & B & gender & $137,806$ & $166,353$ & $44.53$\% & $99.93$\% & $27,562$ ($30$\%) & $229.16$ & $226.97$ & ~\cite{libimsetiDATA}\\
			\noalign{\smallskip}\hline
	\end{tabular}}
\end{table}

For the behavioral data, we build $l2$-regularized Logistic Regression (L2-LR) models and Multi-Layer Perceptrons (MLPs). Logistic Regression has been shown to be the best-performing shallow model for big behavioral data~\cite{2019DeCnuddeBenchmarking}.\footnote{A follow-up study reported a performance improvement by applying deep learning models, at a substantial computational cost~\cite{2019DeCnuddeBenchmarking}} For the textual data, we build bag-of-words support vector machines (SVM) with linear and RBF kernels,
because they are well-established to be successful for text mining applications~\cite{1998JoachimsText,1998DumaisText,2014MartensSEDC}. 
For preprocessing text, we remove stopwords and lemmatize tokens using the \textit{NLTK} package in \textit{Python}, and then, use TF-IDF vectorization~\cite{2014MartensSEDC,1998JoachimsText}.\footnote{TF-IDF is short for term frequency and inverse document frequency.}

Note that, while feature selection and dimensionality reduction techniques could help tackle the high dimensionality of models from textual and behavioral data, previous research has demonstrated that these approaches result in models with worse generalization performance than the ones trained on the original, fine-grained feature space~\cite{2014JunqueBigData,2015ClarkDimReduc,1998JoachimsText,2012LiText}. For this reason, we do not conduct a preprocessing step with dimensionality reduction techniques, but rather build predictive models using the complete feature space to exploit all information captured in the data.

We use $80$\% of the data for training the models and $20$\% as a test set. For L2-LR and SVM, we fine-tune the regularization parameter using a holdout set ($25$\% of training data). For MLP, we use the optimized parameter configuration as found in~\cite{2019DeCnuddeBenchmarking}. We build models using the \textit{Scikit-learn} library.
To make classifications, we set the classification threshold to be approximately equal to the class imbalance.

\subsection{Explanations}
For the experiments, we generate counterfactuals for the positively predicted\footnote{We focus on the positively predicted instances because, most of the time, we are interested in explaining instances predicted as a `class-of-interest' (typically the minority class)~\cite{2014MartensSEDC}. Explaining negatively predicted instances has additional complications, as the negative prediction is the default prediction; see discussion by Martens \& Provost~\cite{2014MartensSEDC}.} test instances and we allow the explanation sets to have up to $30$ features. For \textit{SEDC}, we set the maximum number of iterations to $50$ and we use our own \textit{Python} implementation.\footnote{see https://github.com/yramon/edc} 
For \textit{LIME-C}, we use LimeText explainer\footnote{see https://github.com/marcotcr/lime} for generating the importance-ranked list. Currently, no implementation exists for behavioral data, where a single reference value of zero is used. For this reason, we artificially generated textual data from the behavioral features and use the \textit{CountVectorizer}\footnote{This function converts a collection of text documents to a matrix of token counts and is available from the \textit{Scikit-learn} library.}. We set the complexity parameter $K$ equal to the number of active features~\cite{2016RibeiroLIME} and we set the number of perturbed instances $\tilde{n}$ equal to $5000$~\cite{2016RibeiroLIME,2018NguyenText}. Next, we compute counterfactuals from the ranked feature list based on the \textit{lin-SEDC} algorithm~\cite{2014MartensSEDC}, which is an implementation for finding counterfactuals specifically for linear models.
For \textit{SHAP-C}, we first compute the linear model using the model-agnostic implementation  
with a single reference value (zero), default $l1$-regularization and the identity link function.\footnote{see https://github.com/slundberg/shap} Similar to \textit{LIME-C}, we set the size of the neighborhood $\tilde{n}$ equal to $5000$. Here as well, counterfactuals are computed from the ranked list using \textit{lin-SEDC}.

\subsection{Evaluation criteria}

We define the following set of performance metrics for evaluating counterfactual explanations generated by the three different algorithms:
\begin{enumerate}
	\item \textbf{Effectiveness}
	\begin{itemize}
		\item \textbf{Switching point}: number of features that need to be removed before the classification changes. The switching point equals the size of the counterfactual explanation.
		\item \textbf{Percentage explained}: fraction of positively predicted instances for which a counterfactual explanation smaller than $30$ features is found.
	\end{itemize}
	\item \textbf{Efficiency}
	\begin{itemize}
		\item \textbf{Computation time}: number of seconds it takes to generate an explanation.
	\end{itemize}
\end{enumerate}

To compare the effectiveness of the different algorithms, we need a common definition for assessing (counterfactual) explanations. 
Feature-ranking explanations were tied to the notion of the counterfactual implicitly by Nguyen~\cite{2018NguyenText}, who discussed the ``switching point,'' which is the number of features that need to be deleted---when traversing the ranked list---before the prediction switches to another class.  (This is essentially the procedure of \emph{lin-SEDC}~\cite{2014MartensSEDC}.)  The switching point was originally introduced by Nguyen as a proxy for the method's ability to rank features from high to low relative importance~\cite{2018NguyenText,2016ArrasLRPText}; it also gives us a straightforward method for turning the feature-ranked explanations into counterfactual explanations.
(For explanations already represented as counterfactuals, such as those produced by \textit{SEDC}, the switching point simply equals the number of features in the explanation.)  Measuring the switching point is important, because in cases where the prediction is not the default prediction, simply selecting all the features would produce a class change, but would be a trivial `explanation'. All else being equal, for a better importance-ranked list one would not have to choose as many features to create a counterfactual explanation, resulting in a lower switching point.
We do not allow counterfactuals to be larger than $30$, thus, also the switching point will be no larger than $30$. 

We compute a random explainer for estimating the switching point, against which to compare the counterfactual algorithms. It randomly selects a feature and sets it to zero. If the class changes, then a switching point is found. If not, it verifies whether the predicted score at least decreased. If not, it selects another random feature. If yes, then it selects a new, random feature and evaluates whether leaving out these two features together results in a class change. This is repeated until the random algorithm finds a switching point. Note that we do not restrict the switching point of the random explainer to be smaller than $30$ because else, the subset of instances for which to compare the effectiveness becomes too small. 

Information on effectiveness also is captured by the percentage explained, which indicates the fraction of instances for which a counterfactual explanation smaller than $30$ is found. More specifically, when the explanation method is not very good at identifying the most relevant features, the algorithm will most likely compute larger explanations (larger swiching points). This will result in fewer instances for which a counterfactual smaller than $30$ is found.

We also compare the efficiency of available implementations of the explanation algorithms, as finding counterfactual explanations can be a hard computational problem (cf., \textit{Why complete search fails}, above). The computation time is important to a greater or lesser degree depending on the timeliness needs of the application.  For example, whether one will compute an explanation on demand for a small number of instances at human-cognition speeds \textit{versus} one will compute and cache explanations for all predictions in a high-throughput application (e.g., \textit{why was I shown this?}).

In sum, we evaluate the algorithms on (1) the size of the counterfactual explanations they generate, which we refer to as switching points (smaller-sized explanations are better as they capture the most important features), (2) the percentage of test instances that they have explained, and (3) the computation time they need to generate explanations.


\paragraph{McNemar mid-p significance test.}
For each data set and each model, we evaluate whether the differences in the percentage of instances explained, switching points, and computation times are statistically significant. We use an adjusted version of the exact conditional McNemar test called the McNemar mid-$p$ test~\cite{2013FagerlandMcNemar}, which tests for marginal homogeneity of two dichotomous variables. This test is simple and frequently used for binary matched-pairs data. Several versions of the test exist and we choose the mid-$p$ version because, in small samples, it has shown a good balance between overly conservative exact tests and asymptotic versions of the test that violate the nominal significance level. 
We refer to \textbf{Appendix~\ref{AppendixA}} for additional information about how the test works. 

\vspace*{-4mm}
\section{Results: Effectiveness}
\label{sec:results_effectiveness}
\vspace*{-2mm}

\textbf{Table~\ref{table:PG_switchingpoint}} shows the percentage explained by each of the algorithms. For the linear models, there are very small differences between the methods and for $12$ out of $13$ data sets, \textit{SEDC} is better than or as good as the other methods. For the \textit{Libimseti} data, \textit{LIME-C} and \textit{SHAP-C} find significantly fewer counterfactual explanations than \textit{SEDC}.

For the nonlinear models, however, \textit{SEDC} never produces more explanations than \textit{LIME-C} and \textit{SHAP-C}.  \textit{SEDC} has a significantly lower percentage explained than \textit{LIME-C} and/or \textit{SHAP-C} for $5$ out of $13$ data sets. Since in theory, without an iteration limit, the best-first search will find (all) explanations for every case, this phenomenon is due to a heuristic cut-off of the search at $50$ iterations---it does not expand more than $50$ feature sets (search nodes).  
In more detail: for some nonlinear models, removing one feature does not result in a predicted score change for any of the features. Consequently, the algorithm selects a random feature to continue with in the following iteration.  The same may happen in the second iteration. These `bad' feature choices are what makes the algorithm need more than $50$ iterations to find a counterfactual explanation. 

Furthermore, \textit{SHAP-C} seems to have difficulties for the \textit{Fraud} data. For the linear and nonlinear model respectively, only $81.67$\% and $75$\% of the test instances are explained. The non-explained instances were the instances with more than $12$ active features. For these instances, all estimated importance weights (step $1$ in \textbf{Algorithm~\ref{alg:2}}) are zero, so no linear explanation model was generated. We conjecture this is due to the random sampling procedure, which results in a higher number of required instances $\tilde{n}$. When setting the sample size $\tilde{n}$ to $7000$ (instead of $5000$), the percentage explained increases to a maximum of $100$, indicating that this is the required number of perturbed samples needed to generate explanations. We conjecture that the critical number of perturbed samples increases for large instances for highly unbalanced data like \textit{Fraud} and that this is related to the sampling procedure of \textit{SHAP}.  

\begin{table*} \small
	\centering
	\caption{\textbf{Percentage explained (fraction of positively predicted instances for which a counterfactual smaller than $\textbf{30}$ is found)}. For stochastic \textit{LIME-C/SHAP-C}, these are average percentages over $5$ runs. The best percentages are indicated in bold. The percentages are underlined if a method is significantly worse than the best method on a $1$\% significance level using a McNemar mid-p test~\cite{2013FagerlandMcNemar}.} \label{table:PG_switchingpoint}
	\begin{center}\scalebox{0.8}{
			\begin{tabular}{c|rrr|rrr}
				& & \textbf{Linear} & & & \textbf{Nonlinear} \\
				\hline
				\textbf{Dataset} & \textbf{SEDC} (\%) &\textbf{LIME-C} (\%) & \textbf{SHAP-C} (\%) & \textbf{SEDC} (\%) &\textbf{LIME-C} (\%) & \textbf{SHAP-C} (\%) \\
				\noalign{\smallskip}\hline
				Flickr & $\textbf{100}$ & $\textbf{100}$ & $\textbf{100}$ & $\textbf{28.67}$ & $28.33$ & $\textbf{28.67}$\\
				Ecommerce & $\textbf{100}$ & $97.33$ & $\textbf{100}$ & \underline{$95.00$}  & \underline{$97.00$} & $\textbf{99.67}$ \\
				Airline & $\textbf{100}$ & $\textbf{100}$ & $\textbf{100}$ & $\textbf{100}$ & $\textbf{100}$ & $\textbf{100}$\\
				Twitter & $\textbf{100}$ & $\textbf{100}$ & $\textbf{100}$ & $\textbf{100}$ & $\textbf{100}$ & $\textbf{100}$\\
				Fraud & $\textbf{100}$ & $\textbf{100}$ & \underline{$81.67$} & $\textbf{100}$ & $\textbf{100}$ & \underline{$75$}\\
				YahooMovies & $\textbf{100}$ & $\textbf{100}$ & $\textbf{100}$ & $98.67$ & $\textbf{100}$ & $\textbf{100}$\\
				TaFeng & $\textbf{100}$ & $\textbf{100}$ & $\textbf{100}$ & \underline{$93.33$} & $\textbf{100}$ & $\textbf{100}$\\
				KDD2015 & $\textbf{100}$ & $\textbf{100}$ & $\textbf{100}$ & $99.67$ & $\textbf{100}$ & $99.67$\\
				20news & $99.47$ & $99.47$ & $\textbf{100}$ & $99.47$ & $98.94$ & $\textbf{100}$\\
				Movielens\_100k & $\textbf{100}$ & $\textbf{100}$ & $\textbf{100}$ & $\textbf{100}$ & $\textbf{100}$ & $\textbf{100}$\\
				Facebook & $\textbf{95.67}$ & $95.00$ & $95.00$ & \underline{$70.33$} & $\textbf{92.67}$ & \underline{$89.67$}\\
				Movielens\_1m & $\textbf{98.67}$ & $\textbf{98.67}$ & $\textbf{98.67}$ & \underline{$88.33$} & $95.00$ & $\textbf{95.67}$\\
				Libimseti & $\textbf{92.67}$ & \underline{$90.33$} & \underline{$88.67$} & \underline{$77.00$} & $\textbf{81.67}$ & \underline{$72.33$}\\
				\noalign{\smallskip}\hline
				Average & $\textbf{98.96}$ & $98.52$ & $97.23$ & $88.49$ & $\textbf{91.82}$ & $89.28$\\
				\# wins & $\textbf{12}$ & $9$ & $10$ & $5$ & $\textbf{9}$ & $\textbf{9}$\\
				\noalign{\smallskip}\hline
		\end{tabular}}
	\end{center}
	\vspace*{-5mm}
\end{table*}

\begin{table*}
	\centering
	\caption{Median and interquantile range of \textbf{switching point}. For stochastic \textit{LIME-C/SHAP-C}, this is the average median/range over $5$ runs. The switching point is measured over the subset of instances where \textit{all} methods have found a switching point. The best (median) switching points are indicated in bold. The values are underlined if a method is significantly worse than the best method (smallest median value) on a $1$\% significance level using a McNemar mid-p test~\cite{2013FagerlandMcNemar}.}\label{table:SPbehavioralmedian}
	\begin{center}\scalebox{0.75}{
			\begin{tabular}{c|rrrr|rrrr}
				&  & \textbf{Linear} & & & & \textbf{Nonlinear} \\
				\noalign{\smallskip}\hline
				\textbf{Dataset} & \textbf{SEDC} &\textbf{LIME-C} & \textbf{SHAP-C} & \textbf{Random} & \textbf{SEDC} &\textbf{LIME-C} & \textbf{SHAP-C} & \textbf{Random}\\
				\noalign{\smallskip}\hline
				Flickr & $\textbf{1(1-1)}$ & $\textbf{1(1-1)}$ & $\textbf{1(1-1)}$ &$\textbf{1(1-1)}$ & $\textbf{1(1-1)}$ & $\textbf{1(1-1)}$ & $\textbf{1(1-1)}$ &$\textbf{1(1-2)}$\\
				Ecommerce & $\textbf{1(1-1)}$ & $\textbf{1(1-1)}$ & $\textbf{1(1-1)}$ & $\textbf{1(1-2)}$  & $\textbf{1(1-1)}$ & $\textbf{1(1-1)}$ & $\textbf{1(1-1)}$ & $\textbf{1(1-1)}$\\
				Airline & $\textbf{1(1-2)}$ & $\textbf{1(1-2)}$ & $\textbf{1(1-2)}$ & \underline{$2(1-3)$} &  $\textbf{1(1-1)}$ & $\textbf{1(1-1)}$ & $\textbf{1(1-1)}$ & \underline{$2(1-3)$}\\
				Twitter & $\textbf{2(1-3)}$ & $\textbf{2(1-3)}$ & $\textbf{2(1-3)}$ & \underline{$3(2-5)$} & $\textbf{1(1-1)}$ & $\textbf{1(1-1)}$ & $\textbf{1(1-1)}$ & \underline{$3(2-5.5)$} \\
				Fraud & $\textbf{1(1-1)}$ & $\textbf{1(1-1)}$ & $\textbf{1(1-1)}$ & $\textbf{1(1-1)}$ & $\textbf{1(1-1)}$ & $\textbf{1(1-1)}$ & $\textbf{1(1-1)}$ & $\textbf{1(1-2)}$\\
				YahooMovies & $\textbf{2(1-4)}$ & $\textbf{2(1-4)}$ & $\textbf{2(1-4)}$ & \underline{$4(2-7)$} & $\textbf{1(1-3)}$ &  \underline{$2(1-3)$} & $2(1-3)$ & \underline{$4(2-12)$ } \\
				TaFeng & $\textbf{2(1-4)}$ & $\textbf{2(1-4)}$ & $\textbf{2(1-4)}$ & \underline{$5(3-11)$} & $\textbf{2(1-8)}$ &  $\textbf{2(1-3)}$ & $\textbf{2(1-3.05)}$ & \underline{$6(3-17)$} \\
				KDD2015 & $\textbf{3(1-7)}$ & $\textbf{3(1-7)}$ & $\textbf{3(1-7)}$ & \underline{$8.5(3-17.25)$} & $\textbf{2(1-3)}$ & $\textbf{2(1-3.95)}$ & $\textbf{2(1-4.5)}$ & \underline{$5(2-9)$}\\
				20news & $\textbf{2(1-4)}$ & $\textbf{2(1-4)}$ & $\textbf{2(1-4)}$ & \underline{$11(4-23.5)$} & $\textbf{1(1-3)}$ & $\textbf{1(1-3)}$ & $\textbf{1(1-3)}$ & \underline{$8(3-18)$} \\
				Movielens\_100k & $\textbf{2(1-4)}$ & $\textbf{2(1-4)}$  & $\textbf{2(1-4)}$ & \underline{$5.5(3-10)$ }& $\textbf{2(1-4)}$ & $\textbf{2(1-4)}$ & $\textbf{2(1-4)}$ & \underline{$5(2-9.25)$} \\
				Facebook & $\textbf{3(2-8)}$ & $\textbf{3(2-8)}$ &$\textbf{3(2-8)}$ & \underline{$8(4-20)$} & \underline{$4(1-13)$} & $\textbf{3(1-4.4)}$ & $\textbf{3(1.2-5)}$ & \underline{$9(4.5-19.5)$}\\
				Movielens\_1m & $\textbf{3(2-7)}$ & $\textbf{3(2-7)}$ & $\textbf{3(2-7)}$ & \underline{$9(4-19.25)$} &  $\textbf{3(1-5)}$ & $\textbf{3(1-6)}$ & $\textbf{3(1-6)}$ & \underline{ $7(3-14)$} \\
				Libimseti & $\textbf{3(2-6)}$ & $\textbf{3(2-6.2)}$ & $\textbf{3(2-6.2)}$ & \underline{$29(13-52)$ }&  $\textbf{2(1-5)}$ & \underline{$4.2(1.8-8.8)$} & \underline{$5(2.5-11.2)$} & \underline{$19(8-38.5)$}\\
				\noalign{\smallskip}\hline
				\# wins & $\textbf{13}$ & $\textbf{13}$ & $\textbf{13}$ & $3$ & $\textbf{12}$ & $11$ & $11$ & $3$\\
				\noalign{\smallskip}\hline
		\end{tabular}}
	\end{center}
	\vspace*{-5mm}
\end{table*}

\textbf{Table~\ref{table:SPbehavioralmedian}} shows the median and interquantile range of the switching points.\footnote{Median reported rather than the mean values because switching point only takes positive values and is right-skewed.}
A first observation is that the data sets with large instances, such as \textit{Movielens\_1m} and \textit{Facebook}, have a wider range of switching points (large third quantile value) compared to data sets with small instances such as \textit{Flickr} and \textit{Ecommerce}, where the first quantile, the median and the third quantile are equal to $1$. We also observe that, for linear models, there are no differences in the median switching point between the algorithms. For linear models, in general, the low switching points of \textit{SEDC} are not a surprising result: it is optimal for linear models, i.e. it will always find the minimum-sized subset of features~\cite{2014MartensSEDC}.
Comparing the results of the novel algorithms \textit{LIME-C} and \textit{SHAP-C}, which are approximation methods, against \textit{SEDC}, for linear models they usually find the smallest-sized explanations as well.

For the nonlinear models, however, no method dominates. \textit{LIME-C} and \textit{SHAP-C} perform worse than \textit{SEDC} on the \textit{YahooMovies} and \textit{Libimseti} data sets. For \textit{Movielens\_1m} and \textit{KDD2015}, the hybrid methods have higher third quantile values, indicating that there are more large counterfactuals compared to \textit{SEDC}. In other words, for some instances, the importance-ranked list computed in the first part of \textbf{Algorithm~\ref{alg:2}} is not sufficiently accurate in order to compute the smallest-sized counterfactual. \textit{SEDC} performs worse in terms of median switching points than \textit{LIME-C} and \textit{SHAP-C} on the \textit{Facebook} data and finds relatively more large counterfactuals for the \textit{TaFeng} data (value of third quantile is $8$ compared to $3$).
The \textit{Facebook} data present an interesting case. The third quantile values of of \textit{SEDC}, \textit{LIME-C} and \textit{SHAP-C} for \textit{Facebook/nonlin} are respectively $13$, $4.4$ and $5$, indicating that there are many large counterfactual explanations. The mean switching points of \textit{SEDC}, \textit{LIME-C} and \textit{SHAP-C} for \textit{Facebook/nonlin} are respectively $8.34$, $3.55$ and $4.13$, indicating that there are more outlier values for \textit{SEDC}. 
The reason here is similar to the discussion of the iteration limit above, but there is an additional factor: we stop the search after the first explanation is found. This may be penalizing \textit{SEDC} in terms of explanation length, but giving it an advantage in terms of computational efficiency. 

Finally, when comparing the methods with the random benchmark, we conclude that all approaches are significantly better at pinpointing the most important features, except for the \textit{Ecommerce}, \textit{Flickr} and \textit{Fraud} data, where random performs as good because of the few active features per instance. This is not very surprising, because setting all active features to zero likely results in a class change. When an instance only has $1$ feature, removing this feature immediately changes the predicted class. The random explainer would also return this explanation. For an instance with $2$ features where removing one of these features would result in a class change, the random explainer has $50$\% chance of removing the right feature and finding the smallest-sized explanation. This indicates that the added value of heuristic search algorithms is low (if not negligible) when explaining a set of very small instances compared very large instances. Note that, for the \textit{Fraud} data, the switching point analysis is conducted over the set of instances where all algorithms have found a counterfactual. For the larger data instances (more than $12$ active features), \textit{SHAP-C} failed to compute counterfactuals and thus, only over the subset of these small instances, the switching points are compared. For this reason, over this subset, the random explainer does not perform significantly worse.

\section{Results: Computational efficiency}
\textbf{Table~\ref{table:computationtimesmedian}} summarizes the computation times. 
We observe that the computation time of \textit{SEDC} is generally better compared to \textit{LIME-C} and \textit{SHAP-C}: for all the data and models, the median computation time for \textit{SEDC} is less than $1$ second. 
The interquantile ranges and the mean computation times also inform us about the efficiency of \textit{SEDC}. More specifically, for data with large instances (\textit{Facebook}, \textit{Movielens\_1m} and \textit{Libimseti}), there are many outlier values for computation times compared to \textit{LIME-C} and \textit{SHAP-C}, as indicated by the large third quantile values. This is because \textit{SEDC}'s efficiency mostly depends on the number of features in the explanation and thus, is more sensitive to the switching point than \textit{LIME-C} and \textit{SHAP-C.} 
We observe that for the data with very low switching points (e.g., \textit{YahooMovies}) \textit{SEDC} is very efficient over the entire set of instances: there are not many extreme values. However, for instances that need more features to be removed before a predicted class change is obtained, \textit{SEDC} is slower. These instances are harder to explain by counterfactuals as they, for example, have many active features that contribute to the model prediction (positive evidence). Moreover, in each iteration of the search, \textit{SEDC} evaluates more and more combinations (the number of combinations to be checked increases nonlinearly with the number of iterations).
\textit{Facebook} again gives an interesting case: for
\textit{Facebook/lin}, the \emph{mean} computation times of \textit{SEDC}, \textit{LIME--C} and \textit{SHAP--C} are respectively $3.02$, $0.69$ and $1.14$ seconds, which indicates that for \textit{SEDC} there are outlier values that push the mean value away from the median value of $0.12$ seconds. For \textit{Facebook/nonlin, Libimseti/lin, Libimseti/nonlin, Movielens\_1m/lin, Movielens\_1m/nonlin} something similar happens. This becomes an issue for classification problems where instances are harder to explain with counterfactuals, i.e. more features need to be removed to change the predicted class. For data with small instances or classification problems where data instances are easier to explain with a counterfactual, \textit{SEDC} is the most efficient algorithm. Note that, despite that \textit{Libimseti} has, on average, a smaller switching point than \textit{Facebook}, it still takes much longer to generate counterfactual explanations for the \textit{Libimseti} data. This is because the number of active features is, on average, very large for \textit{Libimseti}, which also plays an important role in determining the computation time. In \textbf{Appendix~\ref{AppendixD}}, the computation times are plotted against the switching points for each data set. These curves clearly illustrate what happens with \textit{SEDC}'s efficiency for instances with large switching points, and that both \textit{LIME-C} and \textit{SHAP-C} are less sensitive to the number of features in the explanation. 

Overall, \textit{LIME-C} and \textit{SHAP-C} have a stable efficiency: the computation time always stays within a range of $0$ to $9$ seconds and it is almost not sensitive to the switching point. (In contrast to SEDC, for which the computation times range from $0$ to $212$ seconds because of its sensitivity to the number of features in the explanation.) The efficiency of \textit{LIME-C} and \textit{SHAP-C} depends mostly on the number of active features of an instance. The results indicate that \textit{SHAP-C}'s efficiency seems more prone to the number of active features of the instance (median and interquantile range have higher values starting from \textit{YahooMovies}). The relation between the computation times and the number of active features for each of the data sets is also shown in \textbf{Appendix~\ref{AppendixE}}. 

Lastly, the algorithms are generally slower for textual data than for behavioral data. We conjecture this is because of the time to evaluate the SVM decision function, which may be higher compared to the L2-LR and MLP decision functions. As an illustration, take the \textit{Facebook} data (behavioral) and \textit{20news} data (textual). Even though the \textit{Facebook} data has more active features per instance and more features in the model ($122,924$ compared to $41,356$), the median time to compute a counterfactual for all three algorithms is higher for the \textit{20news} data than for the \textit{Facebook} data. A possible explanation is that it takes longer to evaluate the decision function \textit{f} of the SVM model. 

\begin{table*} \small
	\centering
	\caption{Median and interquantile range of \textbf{computation time in seconds}. For stochastic \textit{LIME-C/SHAP-C}, this is the average median/range over $5$ runs. The computation time is measured over the subset of instances where \textit{all} methods have found an explanation. The best (median) computation times are indicated in bold. The values are underlined if a method is significantly worse than the best method (smallest median value) on a $1$\% significance level using a McNemar mid-p test~\cite{2013FagerlandMcNemar}.}\label{table:computationtimesmedian}
	\begin{center}\scalebox{0.75}{
			\begin{tabular}{l|ccc|ccc}
				& & \textbf{Linear} & & & \textbf{Nonlinear} \\
				\noalign{\smallskip}\hline
				\textbf{Dataset}& \textbf{SEDC} & \textbf{LIME-C} & \textbf{SHAP-C} & \textbf{SEDC} & \textbf{LIME-C} & \textbf{SHAP-C}\\
				\noalign{\smallskip}\hline
				Flickr & $\textbf{0.01(0.00-0.02)}$ & \underline{$0.34(0.33-0.35)$} & \underline{$0.08(0.08-0.08)$} & $\textbf{0.02(0.00-0.02)}$ & $\underline{0.39(0.39-0.42)}$ & $\underline{0.12(0.09-0.25)}$ \\
				Ecommerce & $\textbf{0.02(0.00-0.02)}$ & \underline{$0.34(0.33-0.36)$} & $\textbf{0.02(0.02-0.03)}$ & $\textbf{0.02(0.00-0.02)}$ & \underline{$0.39(0.38-0.41)$} & \underline{$0.03(0.03-0.03)$} \\
				Airline & $\textbf{0.02(0.02-0.02)}$ & \underline{$0.94(0.81-1.08)$} & \underline{$0.09(0.03-0.60)$} &  $\textbf{0.02(0.02-0.02)}$  & \underline{$1.35(1.17-1.51)$} & \underline{$0.13(0.04-0.82)$}\\
				Twitter & $\textbf{0.03(0.02-0.05)}$ & \underline{$0.61(0.56-0.64)$} & \underline{$0.18(0.06-0.46)$} & $\textbf{0.02(0.01-0.02)}$& \underline{$0.67(0.63-0.69)$} & \underline{$0.15(0.06-0.47)$}\\
				Fraud & $\textbf{0.01(0.00-0.02)}$ & \underline{$0.38(0.36-0.39)$} & \underline{$0.07(0.06-0.08)$}  & $\textbf{0.01(0.01-0.01)}$& \underline{$0.43(0.42-0.44)$} & \underline{$0.09(0.07-0.17)$} \\
				YahooMovies & $\textbf{0.03(0.02-0.08)}$ &  \underline{$0.44(0.43-0.49)$} & \underline{$0.96(0.90-1.00)$} & $\textbf{0.06(0.03-0.20)}$ & \underline{$0.82(0.79-0.85)$} & \underline{$1.35(1.28-1.39)$}\\
				TaFeng & $\textbf{0.05(0.02-0.22)}$ & \underline{$0.50(0.45-0.59)$} & \underline{$1.03(0.97-1.08)$} &  $\textbf{0.04(0.02-0.40)}$ & \underline{$0.51(0.46-0.59)$} & \underline{$1.01(0.95-1.06)$}\\
				KDD2015 & $\textbf{0.11(0.02-0.79)}$ & $\underline{0.52(0.47-0.61)}$ & \underline{$1.04(0.99-1.09)$} & $\textbf{0.14(0.04-0.56)}$ & \underline{$0.84(0.78-0.94)$} & \underline{$1.37(1.31-1.45)$} \\
				20news & $\textbf{0.19(0.05-1.34)}$ & \underline{$3.12(2.09-4.18)$} & \underline{$3.65(2.74-4.49)$} & $\textbf{0.09(0.03-0.68)}$ & \underline{$2.16(1.49-2.95)$} & \underline{$2.53(1.99-3.09)$}\\
				Movielens\_100k & $\textbf{0.06(0.03-0.30)}$ & \underline{$0.49(0.44-0.69)$} & \underline{$0.87(0.83-1.04)$} & $\textbf{0.09(0.04-0.35)}$ & $\underline{0.55(0.50-0.83)}$ & \underline{$1.10(1.02-1.27)$} \\
				Facebook & $\textbf{0.12(0.03-1.17)}$ & \underline{$0.55(0.46-0.75)$} & \underline{$1.11(1.04-1.23)$} & $\textbf{0.19(0.02-2.20)}$ & $\underline{0.51(0.46-0.59)}$ & \underline{$1.06(1.00-1.12)$}  \\
				Movielens\_1m & $\textbf{0.37(0.06-3.09)}$ & \underline{$0.74(0.52-1.21)$} & \underline{$1.21(1.05-1.53)$} & $\textbf{0.39(0.07-1.56)}$  & $\underline{0.76(0.59-1.12)}$ & \underline{$1.29(1.16-1.54)$}\\
				Libimseti & $\textbf{0.36(0.14-2.26)}$ & \underline{$1.07(0.92-1.38)$} & \underline{$1.37(1.27-1.52)$} & $\textbf{0.39(0.09-1.56)}$ & $\underline{1.02(0.91-1.23)}$ & \underline{$1.42(1.35-1.53)$}\\
				\noalign{\smallskip}\hline
				\# wins & $13$ & $0$ & $1$ & $13$ & $0$ & $0$\\
				\noalign{\smallskip}\hline
		\end{tabular}}
	\end{center}
	\vspace*{-5mm}
\end{table*}

\vspace*{-5mm}
\section{Conclusion}
\label{sec:1}
\vspace*{-2mm}
From this study applying two novel and one existing model-agnostic explanation algorithms to the finding of counterfactual explanations for high-dimensional behavioral and textual data, we can draw several conclusions. 

First, the (straightforward) extensions \textit{LIME-C} and \textit{SHAP-C} as expected find reasonable, if not always optimal, counterfactual explanations. Furthermore, extending these algorithms to find counterfactual explanations addresses an open problem with the application of these methods to high-dimensional data, namely, which features should be reported in the explanation. The answer for \textit{LIME-C} and \textit{SHAP-C} is: those that allow the creation of an evidence counterfactual. \textit{SHAP-C} does have problems with highly unbalanced data sets. Despite this, \textit{SHAP-C} may still be preferable when the user is particularly interested in the theoretical interpretation of the importance weights~\cite{2017LundbergLeeSHAP}.

\textit{SEDC}, which was designed to find counterfactual explanations, is generally fast and effective, but not always. In the main results, \textit{SEDC} was clearly the fastest. It is provably optimal for linear models, and also empirically found smaller counterfactuals (median values) for the non-linear models on two data sets. However, for certain instances on certain data sets, \textit{SEDC}'s run time was comparably quite large.  Furthermore, the search stopping criteria were met before \textit{SEDC} found explanations in a non-negligible number of cases. As a best-first search algorithm, there is an effectiveness--efficiency tradeoff that we did not explore comprehensively in this paper. (In theory, \textit{SEDC} will eventually find minimal counterfactuals for all instances, but this may take a long time.)  Our results show that \textit{SEDC} seems to be the best alternative when facing data with small instances (few active features). For such data sets, the limitation of best-first search for nonlinear models will be negligible because the number of iterations (and total computation time) will still remain small.


As a main conclusion, however, \textit{LIME-C} seems the best model-agnostic alternative because of its stable efficiency and effectiveness over \textit{all} data and models. Compared to \textit{SEDC}, it is less sensitive to the switching point, and whereas \textit{SEDC} has failed to find counterfactuals for some instances for nonlinear models, \textit{LIME-C} showed a stable effectiveness over all data and models. Even for very large data instances that require many features to be removed for a class change, \textit{LIME-C} computes counterfactuals relatively fast. Moreover, the results show that \textit{LIME-C}'s efficiency is less sensitive to the number of active features than \textit{SHAP-C}'s efficiency. 

\vspace*{-5mm}
\section{Further research}
\label{sec:1}
\vspace*{-2mm}

A very interesting implication of this work is revealing that there is a good deal of room for more research on this topic. 

First of all, we can extend the performance comparison of the algorithms by using more data sets and training classification models other than logistic regression, multilayer perceptrons and support vector machines. This could further increase the robustness of the results. This is particularly important because we want to evaluate how the algorithm works irrespective of the data and model characteristics.

Second, instead of \textit{LIME-C} and \textit{SHAP-C}, other hybrid algorithms could be created. For example, \textit{LIME} (or \textit{SHAP}) could be run first to fix a search order for an algorithm like \textit{SEDC}. In those cases where \textit{LIME-C} (\textit{SHAP-C}) produces a great explanation, this new hybrid would find it fast. But the algorithm could keep searching, and would be biased towards trying the best features found by \textit{LIME} (\textit{SHAP}) before the other feature combinations, which likely would lead to finding even better explanations faster. Furthermore, we see optimized search algorithms performing quite well for computationally hard problems~\cite{2018JACM}; we conjecture that such algorithms could be applied here to find optimal explanations fast.

Moreover, we have not studied extensively the efficiency--effectiveness tradeoff of the algorithms proposed in this paper. 
For \textit{LIME-C} and \textit{SHAP-C}, the effectiveness also depends on the number of perturbed samples used for randomly generating the sample for generating the feature importances. The higher number of required samples for SHAP-C for the unbalanced data sets confirmed this in our results. A larger number of perturbed samples likely results in a more effective algorithm (as the importance rankings are more accurate)~\cite{2016RibeiroLIME, 2018NguyenText}. However, the efficiency would likely drop because the sampling procedure would take more time. For \textit{SEDC}, a similar tradeoff exists. When we would increase the maximum number of iterations and let the algorithm keep on searching for better solutions even when a first explanation was found, we may find more or potentially better explanations (for example, for those cases when \textit{SEDC} now fails). However, the computational cost would also increase when making this decision. An evaluation of how these algorithmic assumptions and choices would affect the outcomes is a very interesting question to be investigated in a follow-up study. 

Lastly, another future research direction stems from a limitation of the study: we have simplified the goal of the algorithm for computing a counterfactual explanation to finding a small-sized counterfactual as fast as possible. So, for \textit{SEDC}, the algorithm stops looking when it has found an explanation. For \textit{LIME-C}, the entire ranked list is explored up until the $30$th top-ranked feature. In future research, we can explore other objectives of the search algorithm and compare results. For example, we can let the algorithm search for all counterfactual explanations up to a certain size or allow the algorithm to look for all possible explanations within a certain time limit.


\vspace*{-5mm}
\section{Appendix}
\label{sec:1}

\subsection{Appendix A: Heuristic Best-First Search}
\label{AppendixA}
\textbf{Algorithm~\ref{alg:1}} describes the heuristic best-first search proposed by Martens \& Provost~\cite{2014MartensSEDC}. Note that the pseudocode is slightly different because we let the algorithm stop running when a first explanation is found. For this reason, the search tree pruning step is no longer necessary. Also, we have added an additional pruning step, more specifically, we exclude the combinations that have already expanded from the set of possible combinations to expand on. Lastly, we have adjusted the algorithm to be more general, rather than specifically tailored to a document classification task. 

\begin{algorithm}
	\caption{SEDC algorithm (via best-first search)}
	\label{alg:1}
	\begin{algorithmic}[0.6]
		\State \textbf{Input:} 
		\\
		$\textbf{x}_{i}$$=$$(x_{i1},\dots,x_{im})$ \textcolor{gray}{\% Instance to explain, with $m$ active features}
		\\
		$features$$=$$(f_{1},\dots,f_{m})$  \textcolor{gray}{\% Feature names of active features of $\textbf{x}_{i}$}
		\\
		$C_{M}$: $\textbf{x}_{i}$ $\rightarrow$ $\{0,1\}$    \textcolor{gray}{\% Binary classifier $C_{M}$ with scoring function $f_{C_{M}}$}
		\\
		$h$($\textbf{x}'_{i}$) : $\textbf{x}'_{i}$ $=$ $(x'_{i1},\dots,x'_{im})$ $\rightarrow$ $\textbf{x}_{i}$ \textcolor{gray}{\% Function $h$ maps instance to binary representation} 
		\\
		$maxiter$$=$$50$ \textcolor{gray}{\% Maximum number of iterations}
		\\
		$maxf$$=$min($30$, $m$) \textcolor{gray}{\% Maximum number of features in explanation}
		\\
		$maxtime$$=$$5$ \textcolor{gray}{\% Maximum computation time in minutes}
		\\
		$t$ $=$ $0$ \textcolor{gray}{\% Time elapsed in minutes}
		\\
		\State \textbf{Output:}
		\State Explanation $R$ in $R_{list}$ for which $p\_score\_change$ is maximal
		\State\hspace{\algorithmicindent} $c$ = 	$C_{M}$$(\textbf{x}_{i})$ \textcolor{gray}{\% Class predicted by the trained classifier}
		\State\hspace{\algorithmicindent} $p$ = 	$f_{C_{M}}$$(\textbf{x}_{i})$ \textcolor{gray}{\% Corresponding probability or score}
		\State\hspace{\algorithmicindent} $R_{list}$ = 	$\{\}$ \textcolor{gray}{\% List of explanations}
		\State\hspace{\algorithmicindent} $p\_score\_change$ = 	$\{\}$ 
		\State\hspace{\algorithmicindent} $sets\_to\_expand\_on$ = 	$\{\}$ 
		\State\hspace{\algorithmicindent} $P\_sets\_to\_expand\_on$ = 	$\{\}$ 
		\State\hspace{\algorithmicindent} $expanded\_sets$ = 	$\{\}$ 
		\State\hspace{\algorithmicindent} \textbf{for} $j$ $=$ $1$ $\rightarrow$ $m$ \textbf{do}
		\State\hspace{\algorithmicindent}\hspace{\algorithmicindent} $c_{new}$ = $C_{M}$($\textbf{x}_{i}$$\setminus$$x_{ij}$) \textcolor{gray}{\% Class predicted if feature $j$ from $\textbf{x}_{i}$ is set to zero}
		\State\hspace{\algorithmicindent}\hspace{\algorithmicindent} $p_{new}$ = $f_{C_{M}}$($\textbf{x}_{i}$$\setminus$$x_{ij}$) \textcolor{gray}{\% Score predicted if feature $j$ from $\textbf{x}_{i}$ is set to zero}
		\State\hspace{\algorithmicindent}\hspace{\algorithmicindent} \textbf{if} $c_{new}$ $\neq$ $c$ \textbf{then}
		\State\hspace{\algorithmicindent}\hspace{\algorithmicindent}\hspace{\algorithmicindent} $R$ = $R$ $\cup$ $features_{j}$ 
		\State\hspace{\algorithmicindent}\hspace{\algorithmicindent}\hspace{\algorithmicindent} $p\_score\_change$ = $p\_score\_change$  $\cup$ ($p$ - $p_{new}$) 
		\State\hspace{\algorithmicindent}\hspace{\algorithmicindent} \textbf{else}
		\State\hspace{\algorithmicindent}\hspace{\algorithmicindent}\hspace{\algorithmicindent} $sets\_to\_expand\_on$ = 	$sets\_to\_expand\_on$ $\cup$ $features_{j}$ 
		\State\hspace{\algorithmicindent}\hspace{\algorithmicindent}\hspace{\algorithmicindent} $P\_sets\_to\_expand\_on$ = 	 $P\_sets\_to\_expand\_on$ $\cup$ $p_{new}$
		\State\hspace{\algorithmicindent}\hspace{\algorithmicindent} \textbf{end if}
		\State\hspace{\algorithmicindent} \textbf{end for}
		\State\hspace{\algorithmicindent} $it$ $=$ $2$
		\State\hspace{\algorithmicindent} $t$ $=$ $t$ + $t_{elapsed}$  \textcolor{gray}{\% Add extra time elapsed}
		\State\hspace{\algorithmicindent} \textbf{while} $R_{list}$=$\{\}$ \textbf{\&} $it$$\leq $$maxiter$ \textbf{\&} $t$$\leq $$maxtime$ \textbf{do}
		\State\hspace{\algorithmicindent}\hspace{\algorithmicindent} $sets\_to\_expand\_on$ = remove combinations already in $expanded\_sets$ 
		\State\hspace{\algorithmicindent}\hspace{\algorithmicindent} from $sets\_to\_expand\_on$
		\State\hspace{\algorithmicindent}\hspace{\algorithmicindent} $combo$ = feature in $sets\_to\_expand\_on$ for which ($p$ - $P\_sets\_to\_expand\_on$) is 
		\State\hspace{\algorithmicindent}\hspace{\algorithmicindent} maximal and the size is smaller than $maxf$ \textcolor{gray}{\% The best-first feature (set)}
		\State\hspace{\algorithmicindent}\hspace{\algorithmicindent} $expanded\_sets$ = $expanded\_sets$ $\cup$ $combo$
		\State\hspace{\algorithmicindent}\hspace{\algorithmicindent} $combo\_set$ = create all expansions of $combo$ with one feature
		\State\hspace{\algorithmicindent}\hspace{\algorithmicindent} $combo\_set2$ = remove combinations already in $sets\_to\_expand\_on$ from $combo\_set$ 
		\State\hspace{\algorithmicindent}\hspace{\algorithmicindent} \textbf{for} combos $C_{0}$ in $combo\_set2$ \textbf{do}
		\State\hspace{\algorithmicindent}\hspace{\algorithmicindent}\hspace{\algorithmicindent}  $c_{new}$ = $C_{M}$($\textbf{x}_{i}$$\setminus$$C_{0}$) \textcolor{gray}{\% Class predicted if features $C_{0}$ from $\textbf{x}_{i}$ are set to zero}
		\State\hspace{\algorithmicindent}\hspace{\algorithmicindent}\hspace{\algorithmicindent}  $p_{new}$ = $f_{C_{M}}$($\textbf{x}_{i}$$\setminus$$C_{0}$) \textcolor{gray}{\% Score predicted if features $C_{0}$ from $\textbf{x}_{i}$ are set to zero}
		\State\hspace{\algorithmicindent}\hspace{\algorithmicindent}\hspace{\algorithmicindent} \textbf{if} $c_{new}$ $\neq$ $c$ \textbf{then}
		\State\hspace{\algorithmicindent}\hspace{\algorithmicindent}\hspace{\algorithmicindent}\hspace{\algorithmicindent} $R$ = $R$ $\cup$ $C_{0}$
		\State\hspace{\algorithmicindent}\hspace{\algorithmicindent}\hspace{\algorithmicindent}\hspace{\algorithmicindent} $p\_score\_change$ = $p\_score\_change$  $\cup$ ($p$ - $p_{new}$) 
		\State\hspace{\algorithmicindent}\hspace{\algorithmicindent}\hspace{\algorithmicindent} \textbf{else if} $C_{0}$ not in $expanded\_sets$ \textbf{then}
		\State\hspace{\algorithmicindent}\hspace{\algorithmicindent}\hspace{\algorithmicindent}\hspace{\algorithmicindent} $sets\_to\_expand\_on$ = 	$sets\_to\_expand\_on$ $\cup$ $C_{0}$
		\State\hspace{\algorithmicindent}\hspace{\algorithmicindent}\hspace{\algorithmicindent}\hspace{\algorithmicindent} $P\_sets\_to\_expand\_on$ = 	 $P\_sets\_to\_expand\_on$ $\cup$ $p_{new}$
		\State\hspace{\algorithmicindent}\hspace{\algorithmicindent}\hspace{\algorithmicindent} \textbf{end if}
		\State\hspace{\algorithmicindent}\hspace{\algorithmicindent} \textbf{end for}
		\State\hspace{\algorithmicindent}\hspace{\algorithmicindent} $it$ $=$ $it$ + $1$ \textcolor{gray}{\% Add extra iteration}
		\State\hspace{\algorithmicindent}\hspace{\algorithmicindent} $t$ $=$ $t$ + $t_{elapsed}$  \textcolor{gray}{\% Add extra time elapsed}
		\State\hspace{\algorithmicindent} \textbf{end while}
	\end{algorithmic}
\end{algorithm}

\subsection{Appendix B: Novel Hybrid Search Algorithms}
\label{AppendixB}

\textbf{Algorithm~\ref{alg:2}} describes the hybrid search algorithm we have proposed in this paper by combining additive feature attribution techniques LIME~\cite{2016RibeiroLIME} and SHAP~\cite{2017LundbergLeeSHAP} with the model-specific algorithm \textit{lin-SEDC} for computing counterfactuals for linear models~\cite{2014MartensSEDC}.

\begin{algorithm}
	\caption{Additive Feature Attribution + Counterfactuals}
	\label{alg:2}
	\begin{algorithmic}[0.8]
		\State \textbf{Input:} 
		\\
		$\textbf{x}_{i}$$=$$(x_{i1},\dots,x_{im})$ \textcolor{gray}{\% Instance to explain, with $m$ active features}
		\\
		$features$$=$$(feature_{1},\dots,feature_{m})$  \textcolor{gray}{\% Feature names of active features of $\textbf{x}_{i}$}
		\\
		$C_{M}$: $\textbf{x}_{i}$ $\rightarrow$ $\{0,1\}$ \textcolor{gray}{\% Binary classifier $C_{M}$ with scoring function $f_{C_{M}}$}
		\\
		$h$($\textbf{x}'_{i}$) : $\textbf{x}'_{i}$ $=$ $(x'_{i1},\dots,x'_{im})$ $\rightarrow$ $\textbf{x}_{i}$ \textcolor{gray}{\% Function $h$ maps instance to binary representation} 
		\\
		$maxf$$=$min($30$, $m$) \textcolor{gray}{\% Maximum number of features in counterfactual}
		\\
		$maxtime$$=$$5$ \textcolor{gray}{\% Maximum computation time in minutes}
		\\
		\State \textbf{Step 1:} AFA($\textbf{x}_{i}$, $h$, $f_{C_{M}}$) \textcolor{gray}{\% Additive Feature Attribution without complexity control}
		\State \textbf{Output:} 
		\\
		$\phi_{i}$$=$$(\phi_{i1},\dots,\phi_{im})$ \textcolor{gray}{\% Estimated coefficients of linear explanation model}
		\\
		$t$ \textcolor{gray}{\% Time elapsed in minutes}
		\\
		\State \textbf{Step 2:} lin-SEDC($\phi_{i}$, $features$, $maxfeatures$, $maxtime$, $t$) \textcolor{gray}{\% lin-SEDC algorithm}
		\State \textbf{Output:}
		\State Explanation in $R$ 
		\State\hspace{\algorithmicindent} Sort coefficients $\phi_{ij}$ in $\phi_{i}$ in descending order as $1$$\dots$$m$
		\State\hspace{\algorithmicindent} Sort $feature_{j}$ in $features$ according to sorted coefficients vector $\phi_{i}$
		\State\hspace{\algorithmicindent} $c$ = 	$C_{M}$$(\textbf{x}_{i})$ \textcolor{gray}{\% Class predicted by the trained classifier}
		\State\hspace{\algorithmicindent} $p$ = 	$f_{C_{M}}$$(\textbf{x}_{i})$ \textcolor{gray}{\% Corresponding probability or score}
		\State\hspace{\algorithmicindent} $R$ = 	$\{\}$ \textcolor{gray}{\% List of explanations}
		\State\hspace{\algorithmicindent} $p\_score\_change$ = 	$\{\}$
			\State\hspace{\algorithmicindent} $C_{0}$ = 	$\{\}$ \textcolor{gray}{\% List of features}
		\State\hspace{\algorithmicindent} $j$=$1$
		\State\hspace{\algorithmicindent} \textbf{while} $R$ = $\{\}$ \textbf{\&} $j$$\leq $$maxf$ \textbf{\&} $t$$\leq $$maxtime$ \textbf{\&} $\phi_{ij}$ $\geq$ $0$ \textbf{do}
		\State\hspace{\algorithmicindent}\hspace{\algorithmicindent} $C_{0}$ = $C_{0}$ $\cup$ $features_{j}$
		\State\hspace{\algorithmicindent}\hspace{\algorithmicindent} $\textbf{x}_{i}$ = ($\textbf{x}_{i}$$\setminus$$C_{0}$)  \textcolor{gray}{\% Remove (set to zero) features in $C_{0}$ from $\textbf{x}_{i}$}
		\State\hspace{\algorithmicindent}\hspace{\algorithmicindent} $c_{new}$ = $C_{M}$$(\textbf{x}_{i})$ \textcolor{gray}{\% Class predicted if feature $j$ from $\textbf{x}_{i}$ is set to zero}
		\State\hspace{\algorithmicindent}\hspace{\algorithmicindent} $p_{new}$ = $f_{C_{M}}$$(\textbf{x}_{i})$ \textcolor{gray}{\% Score predicted if feature $j$ from $\textbf{x}_{i}$ is set to zero}
		\State\hspace{\algorithmicindent}\hspace{\algorithmicindent} \textbf{if} $c_{new}$ $\neq$ $c$ \textbf{do}
		\State\hspace{\algorithmicindent}\hspace{\algorithmicindent}\hspace{\algorithmicindent} $R$ = $R$ $\cup$ $C_{0}$
		\State\hspace{\algorithmicindent}\hspace{\algorithmicindent} $j$ = $j$ + $1$ \textcolor{gray}{\% Add extra iteration}
		\State\hspace{\algorithmicindent}\hspace{\algorithmicindent} $t$ = $t$ + $t_{elapsed}$ \textcolor{gray}{\% Add extra time elapsed}
		\State\hspace{\algorithmicindent} \textbf{end while}
	\end{algorithmic}
\end{algorithm}

\newpage
\subsection{Appendix C: Details on McNemar mid-p test}
\label{AppendixC}
For each data set, we evaluate whether the differences in the percentage of instances explained, switching points and computation times are statistically significant. We use an adjusted version of the exact conditional McNemar called the McNemar mid-$p$ test~\cite{2013FagerlandMcNemar}, which tests for marginal homogeneity of two dichotomous variables. This test is a simple and frequently-used test for binary matched-pairs data. Several versions of the test exist and we choose the mid-$p$ version because, in small samples, it has shown a good balance between overly conservative exact tests and asymptotic versions of the test that violate the nominal significance level. Exact McNemar tests tend to be overly conservative: they tend to produce unnecessary large $p$ values and have poor power in small samples. Asymptotic tests, on the other hand, may violate the nominal significance level for small sample sizes because required asymptotics do not hold in small samples. Traditionally, statisticians advise to use asymptotic tests in large samples and exact tests in small samples. The mid-$p$ test, however, reaches a good compromise between power and violation the significance level. Moreover, it is very easy to calculate in practice, which makes it a good alternative to the complex, exact unconditional version of the test. 

By using the mid-$p$ test, the test is no longer exact (i.e., guaranteed to have type-\RomanNumeralCaps{1} error\footnote{A type-\RomanNumeralCaps{1} error or false positive error is the rejection of a true null hypothesis.} bounded at the nominal level), but the test has type-\RomanNumeralCaps{1} error rates that are, on average, closer to the nominal level than the exact conditional McNemar test, and thus less conservative. Essentially, exact conditional $p$-values lead to the probability of rejecting the null being less than the nominal level for most parameter values in order to make sure it is never greater than the nominal significance level for any parameter values (i.e., very conservative).

Let $N$ denote the number of matched pairs of binomial events $A$ and $B$, where the possible outcomes are referred to as success ($1$) and failure ($2$). Let ($X_{iA}$,$X_{iB}$) denote the outcome of the $i$th pair, where $X_{iA}$ and $X_{iB}$ respectively are the outcomes for the $i$th observation of event $A$ and $B$. The observed data can be summarized in a contingency table as shown in \textbf{Table~\ref{tab:contingency}}. The null hypothesis $H_{0}$ states that the marginal probabilities of success for the binary matched-pairs events $A$ and $B$ are equal. The marginal probabilities that $X_{iA}$ = $1$ and $X_{iB}$ = $1$ are respectively equal to $p_{1+}$ and $p_{+1}$. The test conditions on the number of discordant pairs $n$ ($n$=$n_{12}$+$n_{21}$). The pairs with different values for the events $A$ and $B$ (e.g., $1$-$0$ or $0$-$1$) are called discordant pairs. Under the null hypothesis, $n_{12}$ is binomially distributed with parameters $n$=$n_{12}$+$n_{21}$ and success probability $p$=$0.5$. The alternative hypothesis $H_{A}$ states that the marginal probability of success of binary event $A$ is larger than that of event $B$, or expressed more formally, that $p_{1+}$$>$$p_{+1}$.

\begin{table}
	\centering
	\pgfplotstabletypeset[
	every head row/.style={%
		before row={\toprule 
			& \multicolumn{2}{c}{A}\\            \cmidrule{2-3}},
		after row=\midrule},
	every last row/.style={after row=\bottomrule},
	columns/B/.style={string type},
	columns/Success/.style={string type},
	columns/Failure/.style={string type},
	columns/Total by B/.style={string type},
	]{
		B       Success Failure {Total by B}
		Success                   $n_{11}$($p_{11}$)           $n_{12}$($p_{12}$)             $n_{1+}$($p_{1+}$)
		Failure                 $n_{21}$($p_{21}$)          $n_{22}$($p_{22}$)               $n_{2+}$($p_{2+}$)
		{Total by A}  $n_{+1}$($p_{+1}$)          $n_{+2}$($p_{+2}$)              $N$($1$)
	}
	\caption{Observed counts and outcome probabilities of a paired $2$x$2$ contigency table.}
	\label{tab:contingency}
\end{table}

To calculate the McNemar mid-$p$-value, we start by calculating the McNemar exact conditional test, and condition on the number of discordant pairs $n$. Let $n$ be the sum of the number of discordant pairs $n_{12}$ and $n_{21}$. The sample size $n$ of the test thus reduces to the number of discordant pairs. The one-sided exact conditional $p$-value equals the probability of at least $n_{12}$ successes out of $n$ binomial trials. We use the binomial cumulative distribution function to calculate this. If $n_{12}$ = $n_{21}$, the $p$-value equals 1. We calculate the one-sided mid-$p$-value by subtracting half the binomial point probability of the observed $n_{12}$ from the exact conditional one-sided $p$-value. We assume here that $n_{12}$ is larger than $n_{21}$. The McNemar mid-$p$-value then equals:
\begin{equation}
p_{mid} = p_{exact} - \frac{1}{2}f(n_{12}|n)
\label{eq:midp} 
\end{equation}

The binomial point probability of at least $n_{12}$ successes out of $n$ binomial trials is the conditional probability, under $H_{0}$, of observing $n_{12}$ given the total number of discordant pairs. For the probability of the observed $n_{12}$, we use the binomial probability density function $f$ where we set the probability of success equal to $0.5$:
\begin{equation}
f(n_{12}|n) = {n \choose n_{12}}(\frac{1}{2})^{n}
\label{eq:pointproba} 
\end{equation}

For the exact one-sided $p$-value $p_{exact}$, the binomial cumulative distribution function is used with success probability equal to $0.5$:
\begin{equation}
p_{exact} = \sum_{x_{12}=0}^{min(n_{12},n_{21})} f(x_{12}|n) 
\label{eq:binomcdf} 
\end{equation}

To illustrate how the McNemar mid-$p$ test works in the context of our experiments, consider the \textit{Facebook} data set and the logistic regression model. Consider the computation time as evaluation metric (note: the test is similar for the other criteria). We want to compare the best explanation algorithm in terms of median computation time against the other explanation algorithms. The median computation time of SEDC for \textit{Facebook/lin} is $0.17$ seconds and is lower than the median computation times of LIME-C and SHAP-C. We want to know whether the LIME-C (and SHAP-C) perform significantly worse in efficiently computing counterfactuals compared to SEDC. In the terminology of the McNemar mid-$p$ test, the algorithms SEDC and LIME-C represent the binary events $A$ and $B$ in \textbf{Table~\ref{tab:contingency}}. On the \textit{Facebook} data level, we have $300$ data instances for which we compute a counterfactual explanation. For $70.33$\% of the instances, a counterfactual explanation was found by all explanation algorithms. In other words, we have $N$=$211$ matched-pairs instances. $N$ is the total number of matched-pairs instances. For the $i$th instance, the pair ($X_{iA}$,$X_{iB}$) will be ($1$,$0$) if the SEDC algorithm computed a counterfactual explanation faster for the instance than the LIME-C algorithm. If the computation time for computing a counterfactual for an instance is equal for LIME-C and SEDC, the value for this pair will be ($0$,$0$). The pairs with different values (e.g., $1$-$0$ or $0$-$1$) are called discordant pairs and are used for computing the mid-$p$-value. The observed counts and outcome probabilities for this example are respectively shown in \textbf{Table~\ref{tab:contingencycounts}} and \textbf{Table~\ref{tab:contingencyprobabilities}}. The null hypothesis for our example states that the marginal probabilities of having the smallest computation time for computing a counterfactual are equal for both explanation algorithms. The alternative hypothesis states that the probability of success for the SEDC algorithm is larger than for the LIME-C algorithm. We choose a one-sided hypothesis test rather than a two-sided test because we have a good guess about which explanation algorithm is better in terms of computation time, based on the median computation time. For this reason, the alternative hypothesis is an inequality in terms of $p_{1+}$$>$$p_{+1}$.

The one-sided exact $p$-value equals $0.037$ and the mid-$p$-value equals $0.032$. The conclusion from the test is that the null hypothesis of equal marginal probabilities of success (i.e., lowest computation time) cannot be rejected on a $1$\% significance level. The same hypothesis test is used for comparing the computation times of SEDC and SHAP-C as well as for the other evaluation criteria (switching point and percentage explained).

\begin{table}
	\centering
	\pgfplotstabletypeset[
	every head row/.style={%
		before row={\toprule 
			& \multicolumn{2}{c}{LIME-C}\\            \cmidrule{2-3}},
		after row=\midrule},
	every last row/.style={after row=\bottomrule},
	columns/SEDC/.style={string type},
	]{
		SEDC       Success Failure {Total by SEDC}
		Success                   0           119             119
		Failure                 92          0               92
		{Total by LIME-C}  92          119              211
	}
	\caption{The observed counts of successes for \textit{SEDC} and \textit{LIME-C} for \textit{Facebook/lin}.}
	\label{tab:contingencycounts}
\end{table}

\begin{table}
	\centering
	\pgfplotstabletypeset[
	every head row/.style={%
		before row={\toprule 
			& \multicolumn{2}{c}{LIME-C}\\            \cmidrule{2-3}},
		after row=\midrule},
	every last row/.style={after row=\bottomrule},
	columns/SEDC/.style={string type},
	]{
		SEDC       Success Failure {Total by SEDC}
		Success                   0           0.56             0.56
		Failure                 0.44          0               0.44
		{Total by LIME-C}  0.44          0.56              1
	}
	\caption{The outcome probabilities of successes for \textit{SEDC} and \textit{LIME-C} for \textit{Facebook/lin}.}
	\label{tab:contingencyprobabilities}
\end{table}

\subsection{Appendix D: Computation time vs switching point}
\label{AppendixD}

\begin{figure}
	\centering
	\scalebox{0.33}{\includegraphics{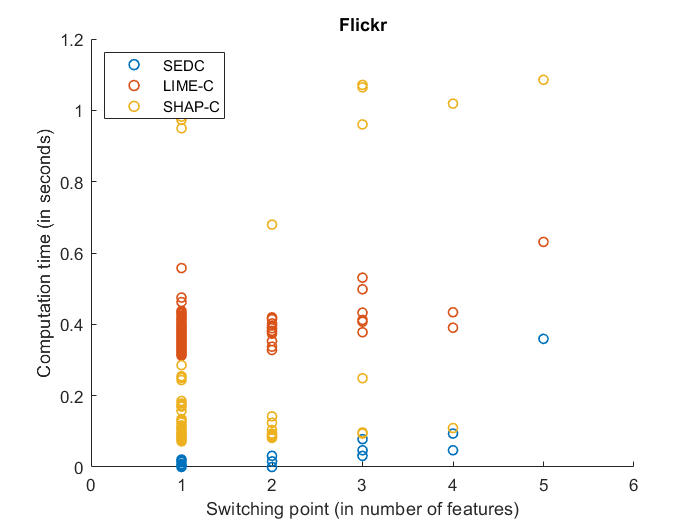}}
	\scalebox{0.33}{\includegraphics{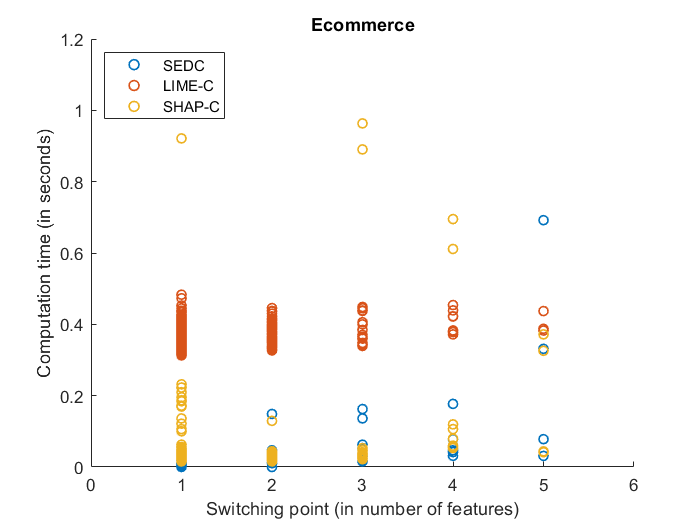}}
	\scalebox{0.33}{\includegraphics{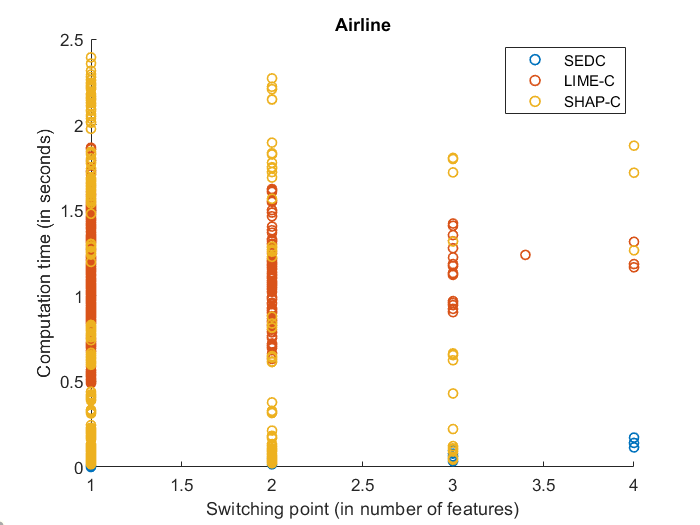}}
	\scalebox{0.33}{\includegraphics{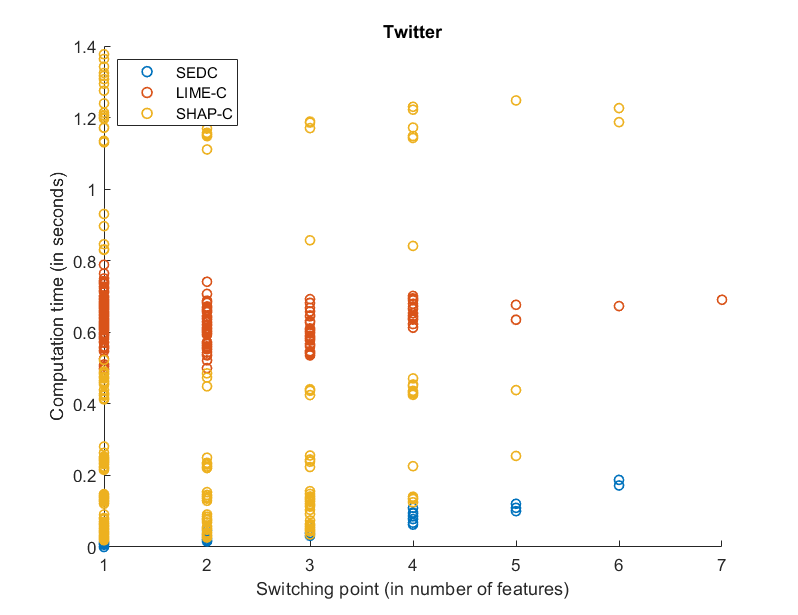}}
	\scalebox{0.33}{\includegraphics{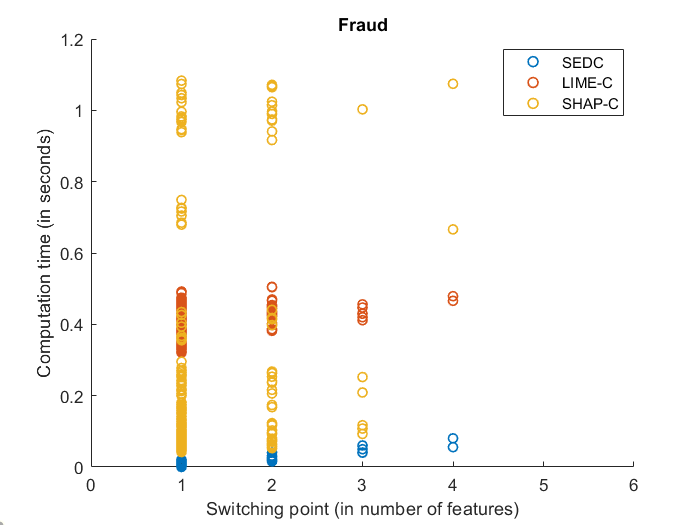}}
	\scalebox{0.33}{\includegraphics{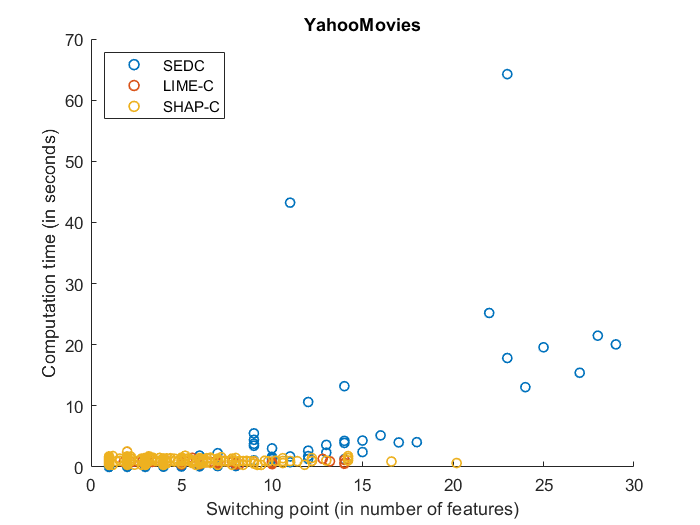}}
	\caption{Comparison of computation times as a function of the switching point for the different algorithms for the data sets \textit{Flickr}, \textit{Ecommerce}, \textit{Airline},  \textit{Twitter}, \textit{Fraud} and \textit{YahooMovies}. Only the instances are plotted for which an explanation was found by \textit{all} the algorithms. Note that we adjusted the scales of the axes for each data set to make sure the results are presented clearly.} \label{CT_SP}
\end{figure}

\begin{figure}
	\centering
	\scalebox{0.33}{\includegraphics{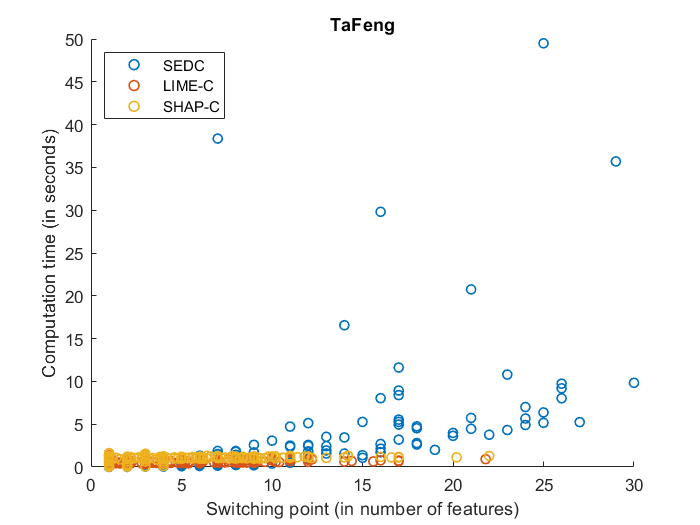}}
	\scalebox{0.33}{\includegraphics{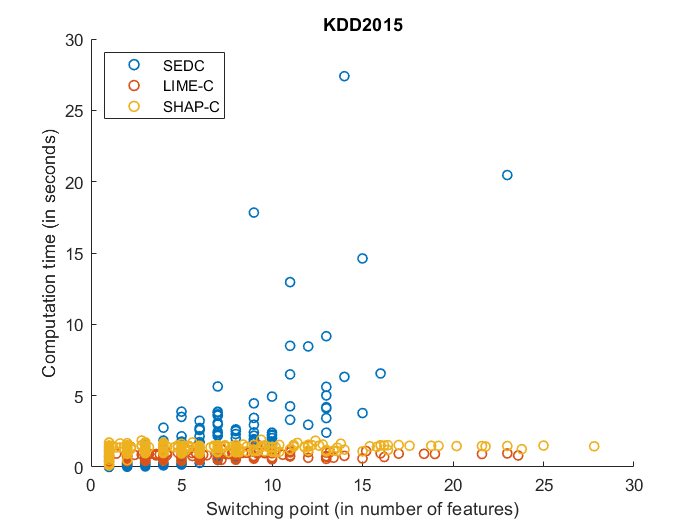}}
	\scalebox{0.33}{\includegraphics{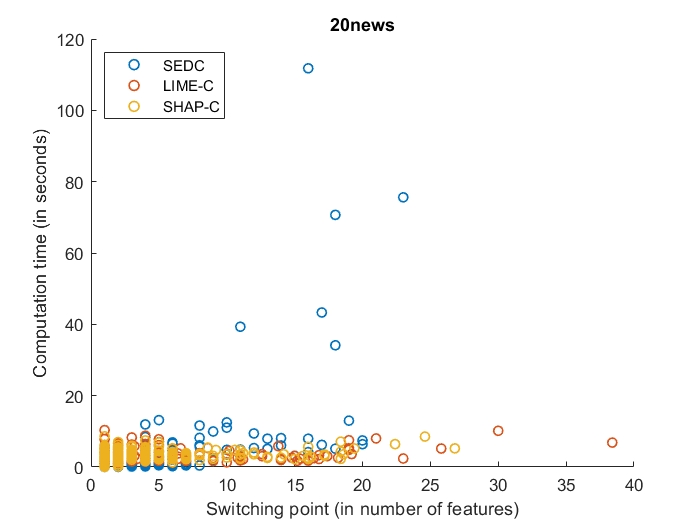}}
	\scalebox{0.33}{\includegraphics{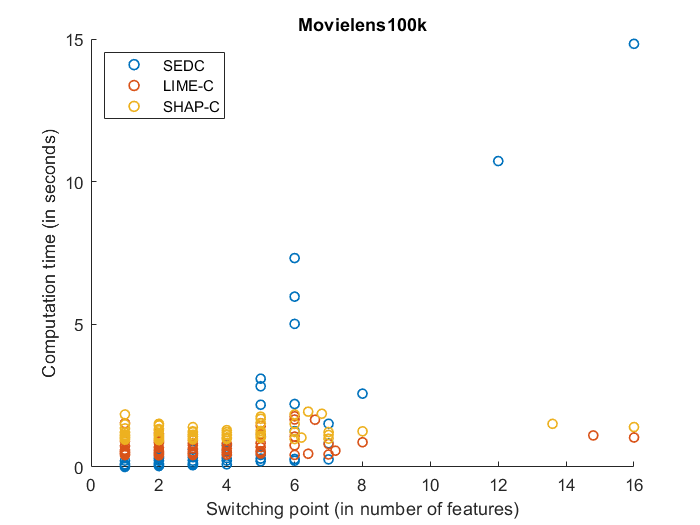}}
	\scalebox{0.33}{\includegraphics{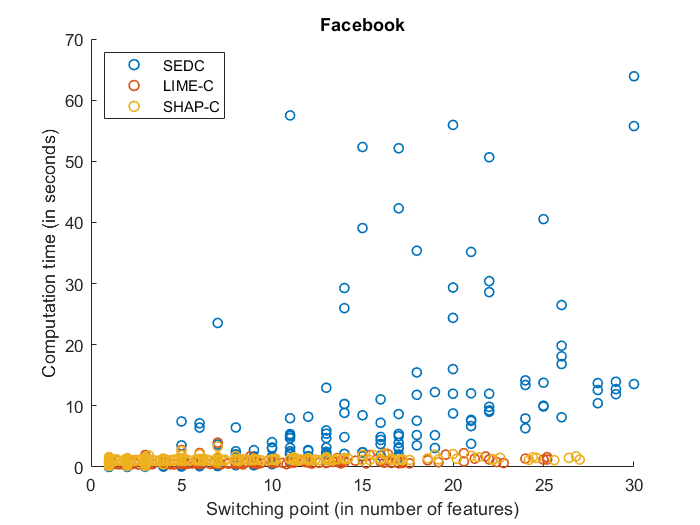}}
	\scalebox{0.33}{\includegraphics{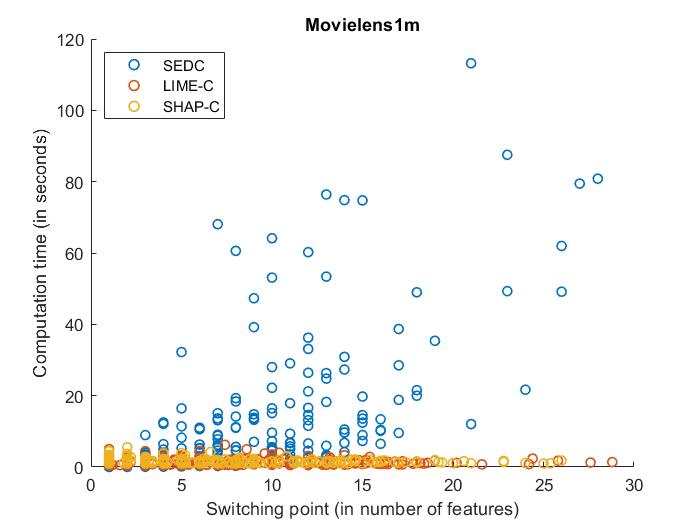}}
	\scalebox{0.33}{\includegraphics{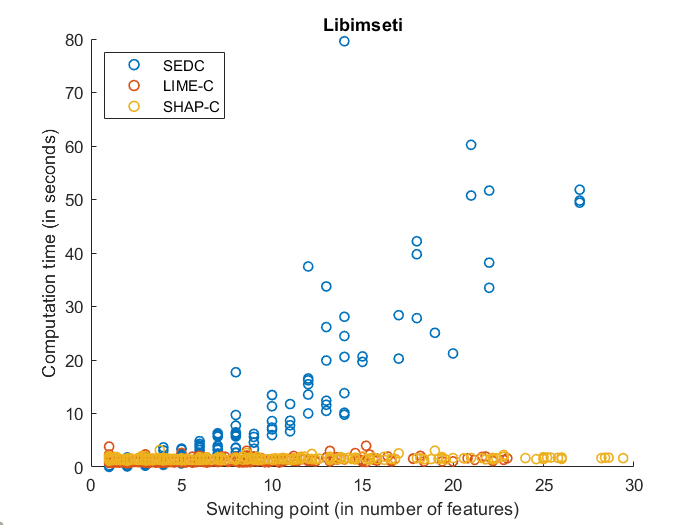}}
	\caption{Comparison of computation times as a function of the switching point for the different algorithms for the data sets \textit{TaFeng, KDD2015, 20news, Movielens\_100k, Facebook, Movielens\_1m} and \textit{Libimseti}. Only the instances are plotted for which an explanation was found by \textit{all} the algorithms. Note that we adjusted the scales of the axes for each data set to make sure the results are presented clearly.} \label{CT_SP2}
\end{figure}

For each data set, we plot the computation times as a function of the switching point (number of features in the counterfactual explanation). Note that we add the positively predicted instances of the linear model and the nonlinear model alltogether in \textbf{Figure~\ref{CT_SP}} and \textbf{Figure~\ref{CT_SP2}}.

From \textbf{Figure~\ref{CT_SP}} we observe that \textit{SEDC} is very efficient for instances that require only a small number of features to be removed before the predicted class changes (switching point). This can be seen for the data sets \textit{Flickr, Ecommerce, Airline, Twitter} and \textit{Fraud}. However, for instances that have larger switching points, the computation time can become very large, especially compared to \textit{LIME-C} and \textit{SHAP-C}. \textbf{Figure~\ref{CT_SP2}} clearly illustrate this phenomenon. Furthermore, \textit{LIME-C}'s efficiency tends to be less sensitive to the switching point compared to \textit{SHAP-C}'s computation times.

\subsection{Appendix E: Computation time vs active features}
\label{AppendixE}

\begin{figure}
	\centering
	\scalebox{0.33}{\includegraphics{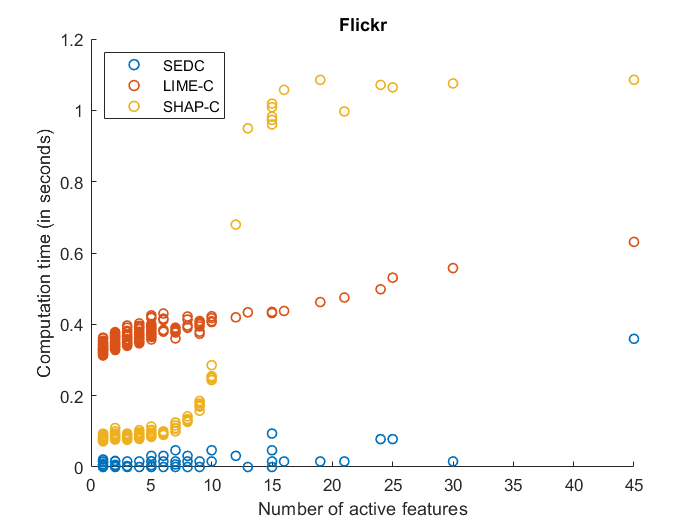}}
	\scalebox{0.33}{\includegraphics{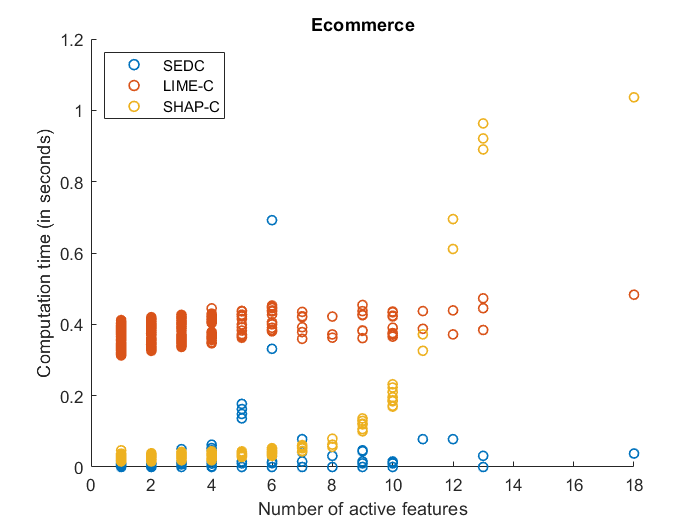}}
	\scalebox{0.33}{\includegraphics{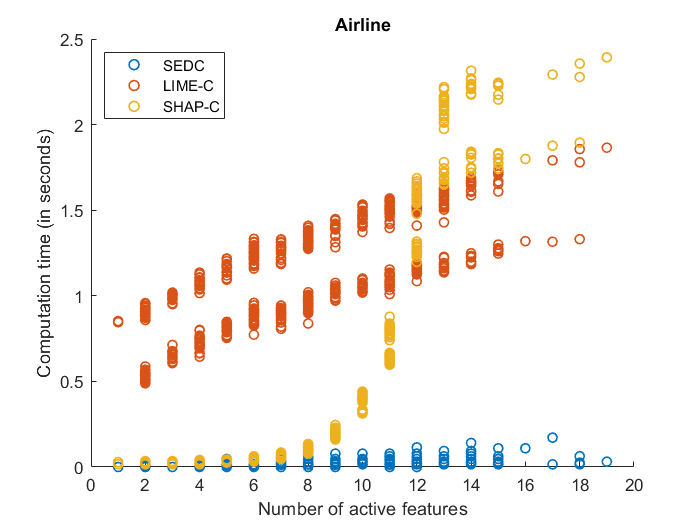}}
	\scalebox{0.33}{\includegraphics{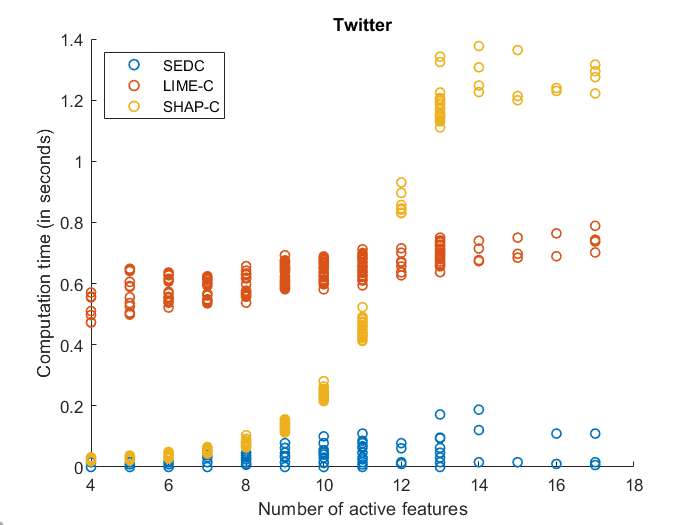}}
	\scalebox{0.33}{\includegraphics{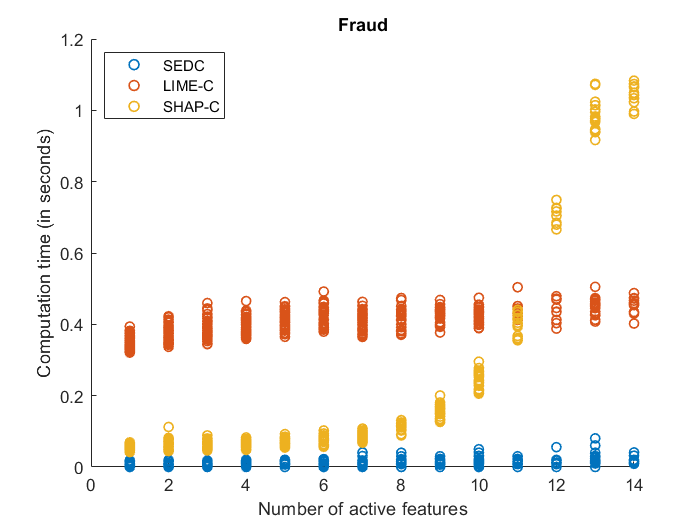}}
	\scalebox{0.33}{\includegraphics{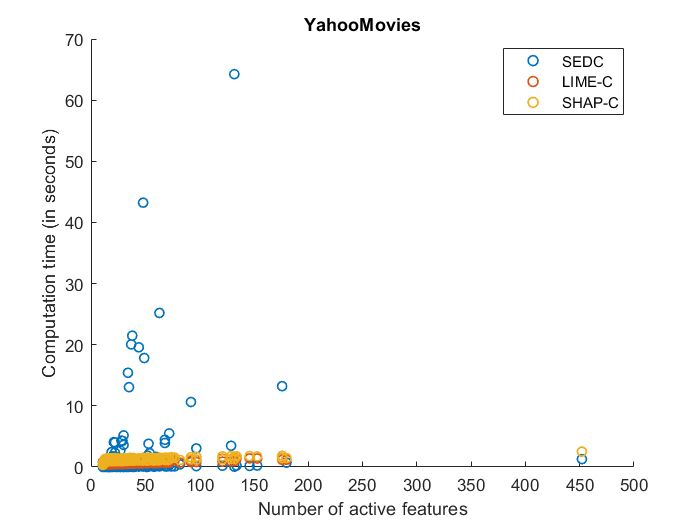}}
	\caption{Comparison of computation times as a function of the number of active features for the different algorithms for the data sets \textit{Flickr}, \textit{Ecommerce}, \textit{Airline},  \textit{Twitter}, \textit{Fraud} and \textit{YahooMovies}. Only the instances are plotted for which an explanation was found by \textit{all} the algorithms. Note that we adjusted the scales of the axes for each data set to make sure the results are presented clearly.} \label{CT_active}
\end{figure}

\begin{figure}
	\centering
	\scalebox{0.33}{\includegraphics{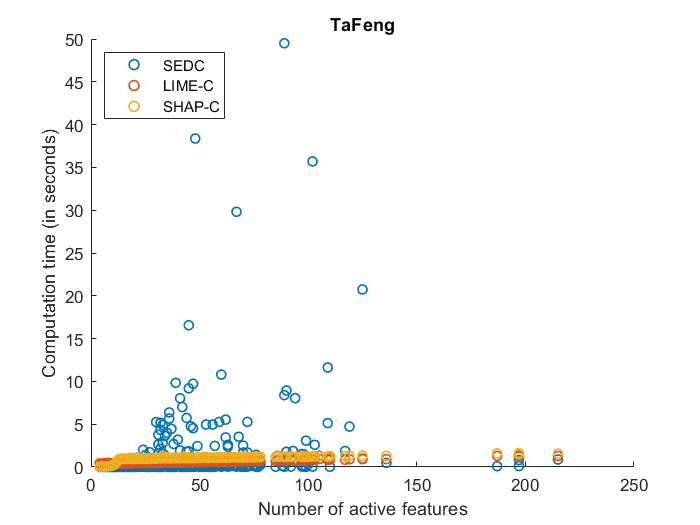}}
	\scalebox{0.33}{\includegraphics{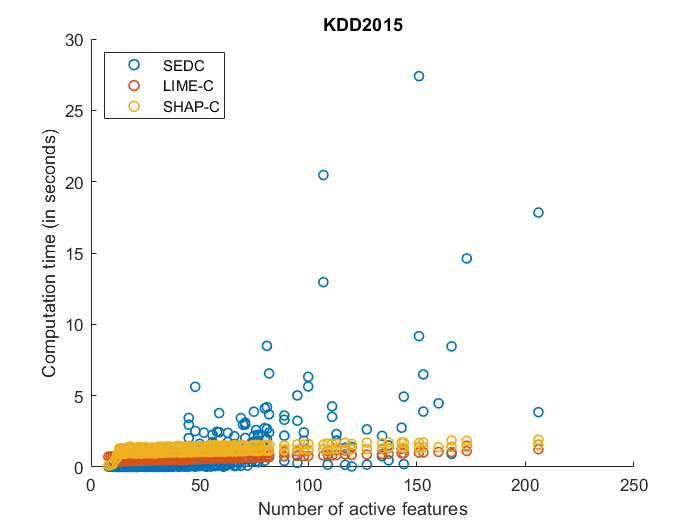}}
	\scalebox{0.33}{\includegraphics{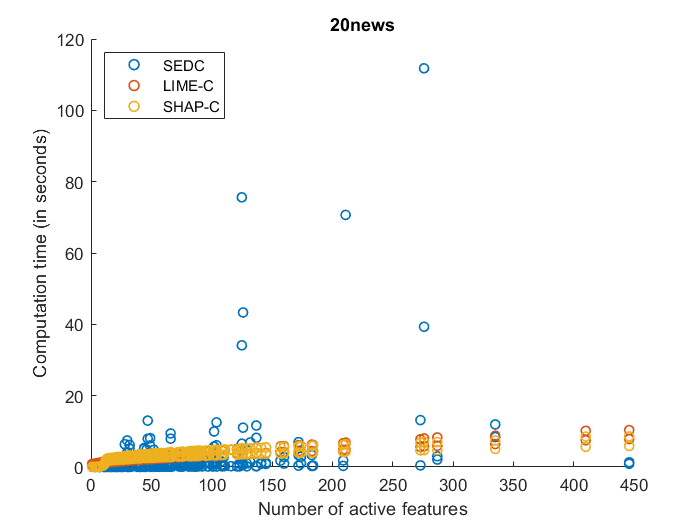}}
	\scalebox{0.33}{\includegraphics{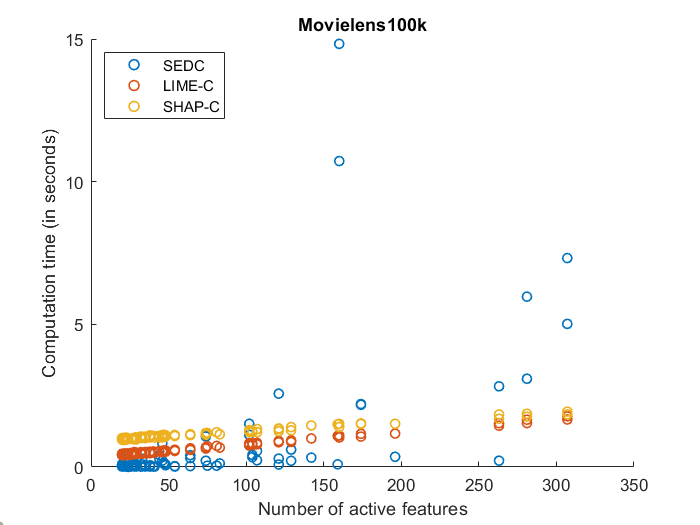}}
	\scalebox{0.33}{\includegraphics{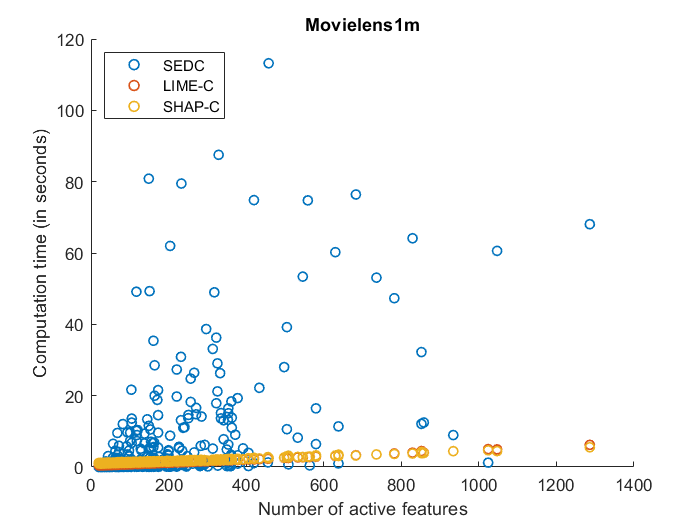}}
	\scalebox{0.33}{\includegraphics{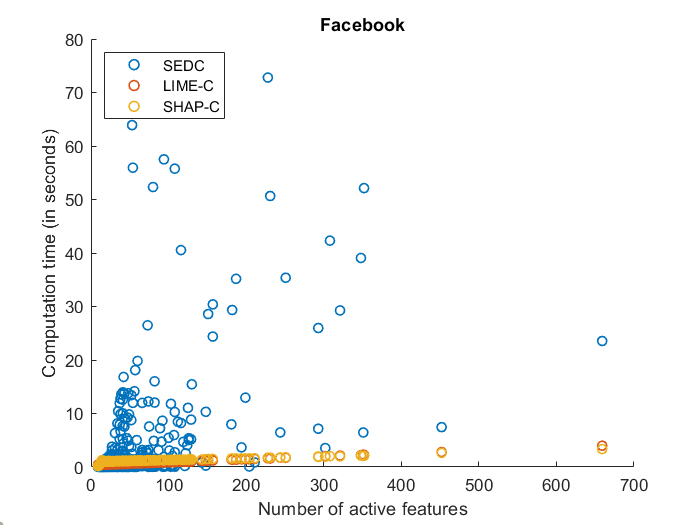}}
	\scalebox{0.33}{\includegraphics{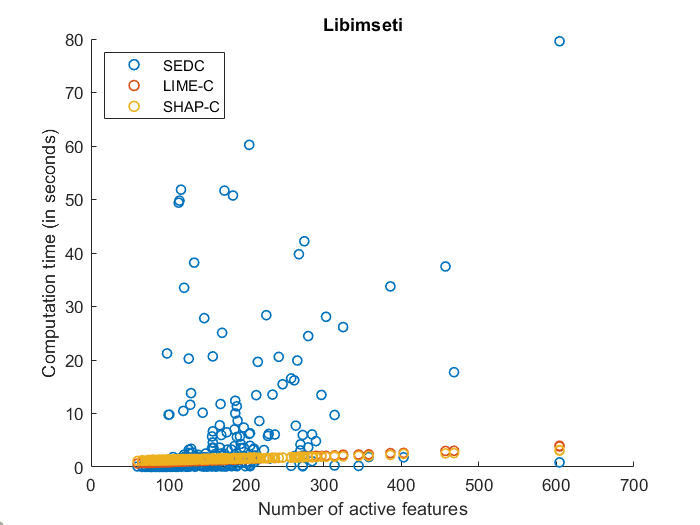}}
	\caption{Comparison of computation times as a function of the number of active features for the different algorithms for the data sets \textit{TaFeng, KDD2015, 20news, Movielens\_100k, Facebook, Movielens\_1m} and \textit{Libimseti}. Only the instances are plotted for which an explanation was found by \textit{all} the algorithms. Note that we adjusted the scales of the axes for each data set to make sure the results are presented clearly.} \label{CT_active2}
\end{figure}

For each data set, we plot the computation times as a function of the number of active features of the instances. Note that we add the positively predicted instances of the linear model and the nonlinear model alltogether in \textbf{Figure~\ref{CT_active}} and \textbf{Figure~\ref{CT_active2}}.

From \textbf{Figure~\ref{CT_active}} and \textbf{Figure~\ref{CT_active2}}, we observe that \textit{LIME-C} has the most stable computation times over the different numbers of active features. In other words, it is least sensitive to the number of active features of the instance to explain. \textit{SHAP-C} seems more sensitive to the number of active features. More specifically, for each data set that has instances larger than $10$ active features, there is a steep (nonlinear) increase between the range of $10$ and $14$. In \textbf{Figure~\ref{CT_active2}}, these level shifts are less obvious as the ranges of number of active features is much wider (up to thousands of active features for \textit{Libimseti}). Nevertheless, \textit{LIME-C} always seems to be less sensitive to the number of active features compared to \textit{SEDC} and \textit{SHAP-C}. Lastly, we observe that for small instances, \textit{SEDC} tends to be more efficient. \textbf{Figure~\ref{CT_active}} depicts that \textit{SEDC} is blazingly fast for very small data instances. However, for larger data instances of data sets like \textit{KDD2015} or \textit{Libimseti}, the computation times of \textit{SEDC} becomes comparably very large. 

\newpage
\bibliographystyle{spmpsci}      
\bibliography{YR_Counterfactuals_WorkingVersion_arxiv}  

\end{document}